\documentclass[10pt,twocolumn,letterpaper]{article}

\usepackage{wacv}              
\usepackage{placeins}

\definecolor{wacvblue}{rgb}{0.21,0.49,0.74}
\usepackage[pagebackref,breaklinks,colorlinks,allcolors=wacvblue]{hyperref}
\usepackage{xcolor}
\usepackage{array}
\usepackage{dblfloatfix}
\definecolor{forestgreen}{rgb}{0.13,0.55,0.13}

\def\wacvPaperID{356} 
\def\confName{WACV}
\def\confYear{2027}

\title{Training-Free Metrics for Synthetic Object Detection Data: \\A Proxy for Detector Performance}

\author{Myeongseok Nam \hspace{0.8cm} Donghun Yeo \hspace{0.8cm} Seungwook Kim \vspace{1.5mm} \\
GenGenAI, South Korea \vspace{1.5mm}\\
\small
}

\begin{document}
\maketitle
\begin{abstract}
Synthetic images are increasingly used to augment scarce real data for object detection.
However, not all synthetic sets help equally, and the only way to know a set's value is to train a detector on it, which is slow and demands dense annotation. 
We ask whether a training-free metric can instead rank candidate synthetic training sets by their downstream utility. 
We propose Conditional-Composition Domain Match (CCDM), which converts any feature-space distance into a composition-stratified comparison, matching candidate and target within metadata-defined strata without training a detector.
On COCO and VisDrone-DET, the best CCDM variant ranks 20 and 19 candidate training sets in strong agreement with YOLOv8 mAP (Spearman $\rho$ = 0.98 and 0.96), outperforming FID, KID, and MMD.
Furthermore, CCDM holds when reference metadata comes from detector pseudo-labels instead of ground truth.
\end{abstract}
\begin{figure*}[!t]
    \centering
    \includegraphics[width=\textwidth]{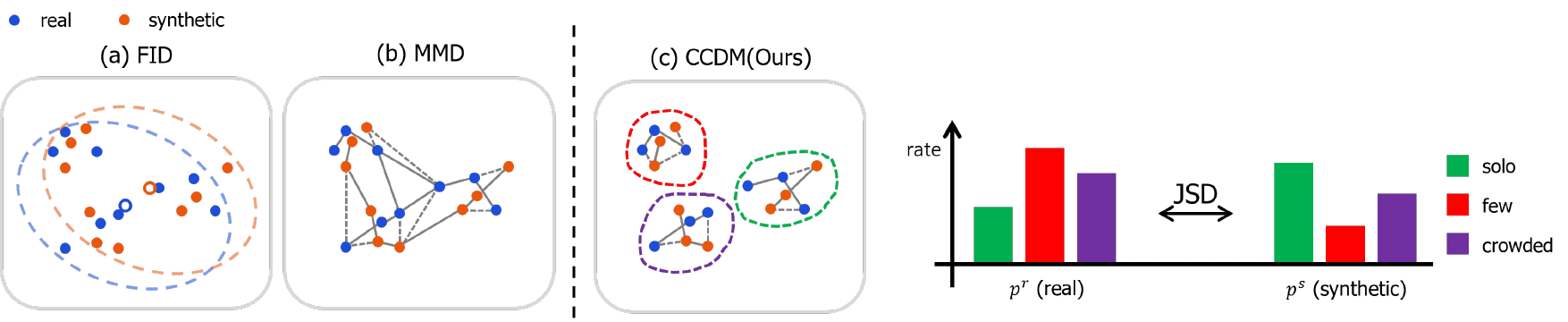}
    \caption{\textbf{Comparison of domain match metrics.} (a) FID fits a single Gaussian to each set and compares global mean and covariance. (b) MMD compares the two distributions globally via pairwise kernel similarities. (c) Our CCDM stratifies images by per-image metadata (\eg, object count: solo, few, crowded), aligns features within each stratum, and measures the mismatch between the metadata compositions $p^r$ and $p^s$ via Jensen-Shannon divergence (JSD).}
    \vspace{-5mm}
    \label{fig:teaser}
\end{figure*} 
\vspace{-6mm}
\section{Introduction}
\label{sec:intro}

Object detection is a fundamental task in computer vision, with architectures advancing from two-stage methods~\cite{girshick2014rich, girshick2015fast, ren2015faster, he2017mask} to single-stage designs such as YOLO~\cite{redmon2016you, redmon2017yolo9000, redmon2018yolov3} and transformer-based models such as DETR~\cite{carion2020end} and RT-DETR~\cite{zhao2024detrs}.
Like all deep learning models, object detection performance depends heavily on the training data~\cite{sun2017revisiting, kaplan2020scaling, deng2009imagenet, liu2022convnet}.
Although large-scale real datasets such as COCO~\cite{lin2014microsoft} exist, data scarcity remains an issue in specialized domains, \eg, where real data is hard to obtain, or for novel object categories or instances where real data do not exist yet~\cite{gupta2019lvis, hu2018learning, kang2019few, wang2020frustratingly, du2019visdrone}.
Even when real data is available, manually annotating bounding boxes is especially resource-intensive.
Recent studies therefore adopt synthetic data for object detection training, since it provides automatic annotations and controllable generation~\cite{feng2024instagen, sankaranarayanan2018learning, gupta2016synthetic}.

However, not all synthetic data is equally useful, and it is unclear which characteristics translate into better detection performance~\cite{feng2024instagen, sankaranarayanan2018learning}.
Currently, the only way to verify a candidate set's value is to train a detector on it and measure the improvement.
This is computationally expensive and time-consuming.
We therefore ask: \textbf{\textit{could a training-free metric serve as a better proxy for the relative utility of a synthetic training set for downstream detection?}}
A natural starting point is the image-set metrics already used to compare generated and real images, such as FID, KID, and MMD.
These reduce the comparison to a single global statistic over a feature distribution, such as the mean and covariance of features~\cite{heusel2017gans, binkowski2018demystifying, salimans2016improved, jayasumana2024rethinking}.
We find that this global view is mis-specified for selecting detection training data, and that it fails in two distinct regimes.
In the first, a candidate set differs from the target in per-image metadata such as object count, box scale, and class mix.
\Cref{fig:bias-axes} gives an example of each.
Global metrics summarize a set with distribution-level statistics, so they respond only weakly to these composition shifts and rank mismatched candidates as close to the target (Section~\ref{sec:preliminary}).
In the second, composition is fixed and candidates differ only in appearance.
Here a single global average pools dissimilar image types together and blurs the fine appearance differences that separate high-mAP pools from low-mAP ones (Section~\ref{sec:experiments}).

Both failures share a cause: the comparison is computed globally rather than conditioned on what each image contains.
As illustrated in \Cref{fig:teaser}, we propose Conditional-Composition Domain Match (CCDM), a family of metrics that converts any classical feature-space distance into a composition-stratified comparison~\cite{heusel2017gans, JMLR:v13:gretton12a}.
CCDM partitions candidate and target into strata defined by per-image metadata, measures appearance agreement within each stratum, and adds a composition penalty scaled by the global distance so the score stays comparable across base metrics (Section~\ref{sec:method}).
Unlike existing training-free metrics~\cite{heusel2017gans, binkowski2018demystifying, JMLR:v13:gretton12a, jayasumana2024rethinking}, which are blind to per-image metadata, CCDM treats metadata as a first-class signal and needs no detector training.
We validate on COCO~\cite{lin2014microsoft} and VisDrone-DET~\cite{du2019visdrone} with YOLOv8 as the downstream detector.
The best CCDM variant ranks 20 COCO candidates (2 real, 18 synthetic) and 19 VisDrone-DET candidates (1 real, 18 synthetic) in strong agreement with downstream mAP (Spearman $\rho = 0.98$ and $0.96$), outperforming classical baselines such as FID, KID, and raw MMD (Section~\ref{sec:experiments}).
The ranking holds even when the real reference set is unlabeled and its metadata comes from off-the-shelf detector pseudo-labels.

Our contributions are summarized as follows:
\begin{itemize}
\item We identify two failure modes of classical training-free metrics (FID, KID, MMD, SWD, and variants) as proxies for synthetic detection data. 
\item We propose Conditional-Composition Domain Match (CCDM), which turns any feature-space distance into a composition-stratified metric to be used as a proxy.
\item On COCO and VisDrone-DET with YOLOv8, a CCDM variant beats every classical baseline. The ranking also survives when reference metadata comes from detector pseudo-labels instead of ground-truth boxes.
\end{itemize}
\vspace{-1mm}
\section{Related works}
\label{sec:related_work}
\vspace{-1mm}
\subsection{Object detection}
\vspace{-1mm}
Object detection~\cite{zou2023object} architectures have evolved from two-stage methods~\cite{girshick2014rich, girshick2015fast, ren2015faster} to single-stage designs~\cite{lin2017focal, liu2016ssd} including the YOLO family~\cite{redmon2016you, redmon2017yolo9000, redmon2018yolov3, bochkovskiy2020yolov4}, and more recently to transformer-based models~\cite{zhu2020deformable, zhang2022dino, meng2021conditional} such as DETR~\cite{carion2020end} and RT-DETR~\cite{zhao2024detrs}.
Across these architectures, detection performance depends heavily on the training data~\cite{sun2017revisiting}.
Benchmarks such as COCO~\cite{lin2014microsoft} and VisDrone-DET~\cite{du2019visdrone} have driven much of this progress, yet collecting and annotating detection data remains costly, which motivates the use of synthetic data for large-scale training~\cite{feng2024instagen}.
We use object detection as the downstream task to validate our metric, which estimates the effectiveness of a synthetic training set without training a detector.

\subsection{Synthetic datasets for computer vision}
\vspace{-1mm}
Synthetic datasets have been adopted across computer vision tasks, including object detection~\cite{feng2024instagen}, semantic segmentation~\cite{richter2016playing, scucchia2025gaming}, and autonomous driving~\cite{johnson2016driving, gaidon2016virtual, dosovitskiy2017carla}.
They provide automatic annotations and controllable generation~\cite{kim2024enhancing, kim2024multi, kim2025rapidmv, kim2025freeaction} that real-world datasets cannot offer, and serve as an alternative when real data is scarce or expensive to collect.
Despite these benefits, synthetic data often exhibits a domain gap from real data~\cite{wang2024improving, sankaranarayanan2018learning, hoffman2018cycada}.
This gap spans two kinds of difference: appearance, and composition such as layout and the distribution of object count, scale, and class.
Either kind can degrade downstream performance, and synthetic data does not always improve it~\cite{wang2024improving, newell2020useful, mayer2018makes}.
Moreover, it is not trivial to identify which characteristics of a synthetic set contribute to downstream performance.
We therefore aim to assess the efficacy of synthetic data for downstream training before training the detector, by proposing a family of metrics dubbed CCDM.
\FloatBarrier
\begin{figure}[t]
  \centering
  \includegraphics[width=\columnwidth]{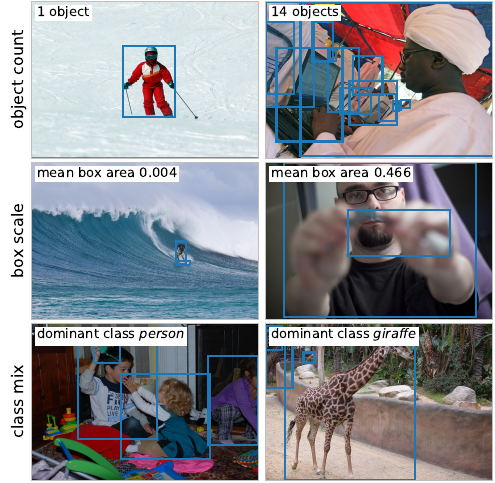}
  \vspace{-6mm}
  \caption{The three metadata axes on real COCO reference images, with ground-truth boxes and each panel's axis value inset. Each pair varies one axis while the other two are held nearly fixed, so the contrast is the axis rather than two unrelated scenes.}
  \vspace{-6mm}
  \label{fig:bias-axes}
\end{figure}

\begin{table}[t]
\centering
\caption{Mean rank (over four target sizes) assigned to three reference candidate classes by eight representative classical metrics on VisDrone-MOT.
Reference groups: \textbf{In-domain matched} = two VisDrone-MOT candidate sets with
balanced composition. \textbf{Cross-domain matched} = three UAVDT candidate sets with balanced
composition. \textbf{In-domain collapsed} = one VisDrone-MOT candidate set with collapsed
composition.
A metric is \emph{appearance-biased} if the in-domain collapsed candidate ranks ahead of the cross-domain matched candidate; \emph{distribution-biased} if it demotes the in-domain collapsed candidate but ranks the cross-domain matched candidate ahead of the in-domain matched; \emph{balanced} if it orders all three correctly, in-domain matched ahead of cross-domain matched ahead of in-domain collapsed. A lower rank means the metric judges that candidate class to be closer to the target, so an unbiased metric should give the two matched classes lower ranks than the collapsed class.}
\vspace{-1mm}
\label{tab:metric-rankings}
\setlength{\tabcolsep}{4pt}
\resizebox{\columnwidth}{!}{%
\begin{tabular}{@{}lccc@{}}
\toprule
Metric &
\shortstack{In-domain\\matched} &
\shortstack{Cross-domain\\matched} &
\shortstack{In-domain\\collapsed} \\
\midrule
\multicolumn{4}{@{}l}{\emph{Appearance-biased}} \\
\midrule
FID                       & 2.75 & 12.50 & 6.50 \\
KID                       & 2.50 & 10.75 & 6.75 \\
FCD\textsubscript{CLIP}   & 2.75 & 12.75 & 6.50 \\
SWD\textsubscript{CLIP}   & 3.00 & 12.25 & 6.00 \\
\midrule
\multicolumn{4}{@{}l}{\emph{Distribution-biased}} \\
\midrule
MMD-RBF\textsubscript{CLIP}            & 5.50 & 5.25 & 13.25 \\
MMD-RBF-multiscale\textsubscript{CLIP} & 5.50 & 5.25 & 13.50 \\
GW\textsubscript{CLIP}                 & 6.75 & 1.50 & 10.75 \\
\addlinespace[2pt]
\midrule
\multicolumn{4}{@{}l}{\emph{Balanced}} \\
\midrule
GW\textsubscript{DINO}              & 5.00 & 5.25 & 13.25 \\
\bottomrule
\end{tabular}%
}
\vspace{-6mm}
\end{table}

\subsection{Metrics for evaluating synthetic image quality}
\vspace{-1mm}
Many metrics have been proposed to compare real and generated image distributions~\cite{heusel2017gans, salimans2016improved, borji2019pros, jayasumana2024rethinking}.
FID~\cite{heusel2017gans} compares the mean and covariance of Inception features under a Gaussian assumption, while KID~\cite{binkowski2018demystifying} replaces the Fréchet distance with an unbiased MMD~\cite{JMLR:v13:gretton12a} estimator, removing that assumption.
Recent variants swap Inception for foundation-model embeddings, \eg CMMD~\cite{jayasumana2024rethinking} (MMD over CLIP) and FCD (Fréchet distance over CLIP).
Image-level and preference-based metrics, such as CLIP-Score~\cite{hessel2021clipscore}, GenEval~\cite{ghosh2023geneval}, PickScore~\cite{kirstain2023pick}, HPSv2~\cite{wu2023human}, ImageReward~\cite{xu2023imagereward}, and SOLACE~\cite{kim2026improving}, instead target text-image alignment or human-judged fidelity.
A common thread is that these metrics judge generated images by a single global comparison and target image fidelity rather than downstream task utility~\cite{jayasumana2024rethinking}.
For object detection, this global view overlooks per-image composition such as object count, scale, and class mix, making it mis-specified for ranking synthetic training data (Section~\ref{sec:preliminary}).
To address this, we propose CCDM, a dataset-level metric that compares candidate and target within metadata-defined strata and adds an explicit composition penalty, so that both appearance and composition are reflected in a single score.
\vspace{-1mm}

\section{Bias in Classical Training-Free Metrics}
\label{sec:preliminary}
\vspace{-2mm}
Classical training-free metrics for image-set comparison (FID~\cite{heusel2017gans}, KID~\cite{binkowski2018demystifying}, MMD~\cite{JMLR:v13:gretton12a}, and their feature-space variants~\cite{jayasumana2024rethinking, radford2021learning, oquab2023dinov2, lee2021learning, lee2023learning}) are commonly used as proxies for downstream training utility, that is, for the performance a detector reaches when trained on the set.
We find this use unreliable for object detection, and isolate why with two observations.
First, even when two candidate sets contain the same mix of scenes, that is the same per-image object counts, box scales, and classes, and differ only in appearance, the differences that move detector mAP are subtle and easily averaged away by a global distance (see the qualitative comparison in the supplementary).
Second, when candidates differ in composition, classical metrics misrank them in three systematic ways, which we name and measure below.
We demonstrate the second point with the bias test below, which motivates the construction of \Cref{sec:method}.
\vspace{-2mm}
\subsection{Bias Test: A Ranking with a Known Answer}
\label{sec:prelim:setup}
\vspace{-1mm}
We build 20 candidate sets of $10{,}000$ aerial frames each from
VisDrone-MOT~\cite{du2019visdrone} and UAVDT~\cite{du2018unmanned}, and score them with $16$
classical training-free metrics against real VisDrone-MOT reference sets of four sizes,
$10$, $100$, $1{,}000$, and $10{,}000$ images.
We sample VisDrone frames from the VisDrone-MOT video sequences, which provide more frames than VisDrone-DET.
Six of the 20 form the three groups the test turns on:
\textbf{in-domain matched} (two sets from the same footage as the reference),
\textbf{cross-domain matched} (three sets from elsewhere),
and \textbf{in-domain collapsed} (one set from the same footage, narrowed to a
single dominant class, sequence, or scale range).
The matched sets are uniform random samples of their source, so nothing skews their
composition, which we call \emph{balanced}. The other 14 are further
composition-shifted sets that fill out the ranking.
For each metric we report a group's mean rank over the four reference sizes, where rank $1$
is the candidate judged closest to the reference set and a group with several members takes
its best member's rank.
A metric that tracks training utility should place both matched groups ahead of the
collapsed one, and the in-domain matched group ahead of the cross-domain matched one.

\begin{figure*}
    \centering
    \includegraphics[width=\textwidth]{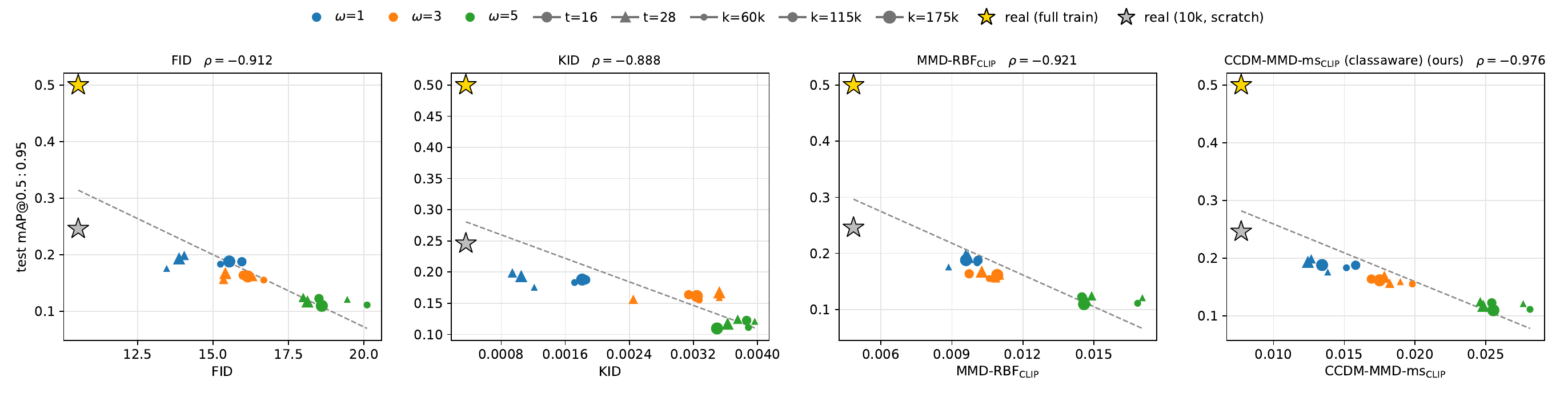}
    \vspace{-8mm}
    \caption{YOLOv8m val2017 mAP@$0.5{:}0.95$ versus four training-free metrics for the 20 COCO candidates of \Cref{tab:syn-map}. The pools share composition, so classical metrics already rank well (FID $0.912$, KID $0.888$, MMD-RBF$_{\mathrm{CLIP}}$ $0.921$). Our CCDM-MMD-RBF-multiscale$_{\mathrm{CLIP}}$ is strongest ($\rho=0.976$). Color: guidance $\omega$. Marker: sampling steps. Size: checkpoint iteration. Stars: the two real candidates, full-\texttt{train2017} in gold and the 10k retrain in gray. Dashed lines: linear fit. Per-panel $\rho$ is signed Spearman (\Cref{sec:prelim:polarity}).}
    \vspace{-4mm}
    \label{fig:metric-vs-map-coco}
\end{figure*}
\begin{figure*}
  \centering
  \includegraphics[width=\textwidth]{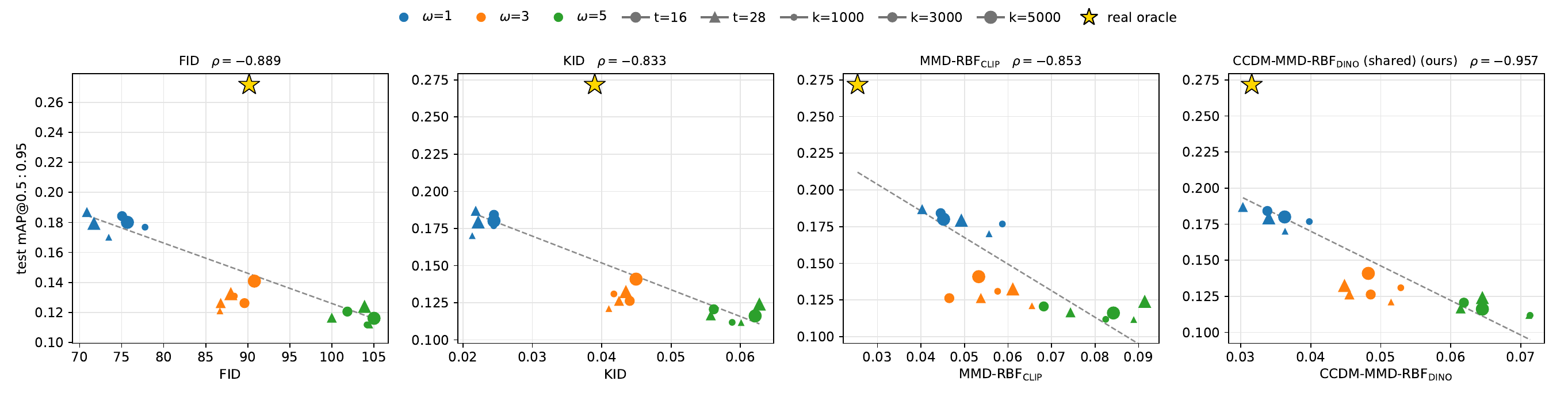}
  \vspace{-8mm}
  \caption{As \Cref{fig:metric-vs-map-coco}, for the 19 VisDrone-DET candidates of \Cref{tab:syn-map}. Unlike COCO, the classical metrics are noticeably weaker here: FID and KID are appearance-biased ($\rho=0.889$, $0.833$) and MMD-RBF$_{\mathrm{CLIP}}$ ($0.853$) makes ranking errors within the $\omega{=}1$ cluster. Our CCDM-MMD-RBF$_{\mathrm{DINO}}$ is strongest ($\rho=0.957$). Encoding and conventions as in \Cref{fig:metric-vs-map-coco}, with a single real candidate here.}
  \vspace{-6mm}
  \label{fig:metric-vs-map-sidebyside}
\end{figure*}
\vspace{-2mm}
\subsection{Three Bias Categories}
\label{sec:prelim:buckets}
\vspace{-1mm}

\Cref{tab:metric-rankings} shows the base metrics grouped by bias category.
\smallbreak
\noindent
\textbf{Appearance-biased metrics.}
These metrics rank the in-domain collapsed candidate ahead of the cross-domain matched candidate, in effect rewarding a candidate for coming from the same footage as the reference, even when its composition is wrong.
Every Fr\'echet-style distance (FID, FCD, DINO-FD), KID, and every transport or moment-matching distance (SWD, CMD, Wasserstein) on CLIP and DINOv2 features falls here.
\smallbreak
\noindent
\textbf{Distribution-biased metrics.}
MMD-RBF variants and Gromov-Wasserstein on CLIP features (GW\textsubscript{CLIP}) correctly place the in-domain collapsed candidate last of the three, at mean rank about $13$ against about $5$ for the cross-domain matched one, but they then rank the cross-domain matched candidate ahead of the in-domain matched one, which is the wrong order.
They react to a mismatch in the overall feature distribution, but not to a composition shift that leaves that overall distribution nearly unchanged.
\smallbreak
\vspace{-2mm}
\noindent
\textbf{Balanced metrics.}
GW\textsubscript{DINO}, the same distance on DINOv2 features, keeps the in-domain matched candidate ahead of the cross-domain matched one while placing the collapsed candidate last, at rank $13.25$.
Since the in-domain matched candidates come from the same footage as the reference and are left unbiased, they match it in both appearance and composition and should rank well ahead of the cross-domain ones, yet even GW\textsubscript{DINO}, the classical metric that comes closest to a correct ordering in this test, separates the two by only 0.25 ranks.
So no classical metric we tested keeps the in-domain matched candidate clearly ahead of both alternatives, which motivates CCDM (\Cref{sec:method}).
\vspace{-2mm}
\subsection{Sign convention for correlations}
\label{sec:prelim:polarity}
\vspace{-2mm}
Every metric we report is a distance or a divergence, so a smaller value means the candidate set looks closer to the reference set.
Subsequent sections report \textbf{signed Spearman} $\rho$, negated so that a positive $\rho$ always means the metric agrees with mAP, that is, the candidates it calls closest to the reference set are the ones that train the best detectors. A negative $\rho$ marks a metric that does the opposite (\eg, several Wasserstein distances in \Cref{tab:syn-ranking}, where greater distance happens to track better mAP).

\section{Method}
\label{sec:method}
\vspace{-1mm}
The bias test of \Cref{sec:preliminary} establishes that only one classical feature-space distance achieves the balanced ordering, and it separates the two matched candidates by just $0.25$ ranks. Every other metric we tested is appearance-biased or distribution-biased.
The appearance and distribution signals that these metrics carry remain useful. What is missing is sensitivity to composition.
The Conditional-Composition Domain Match (CCDM) introduced below adds this missing piece: it computes any feature-space distance separately within groups of images that share the same composition, using only per-image metadata such as object count, box scale, and class mix.

\begin{figure*}[t]
    \centering
    \includegraphics[width=\textwidth]{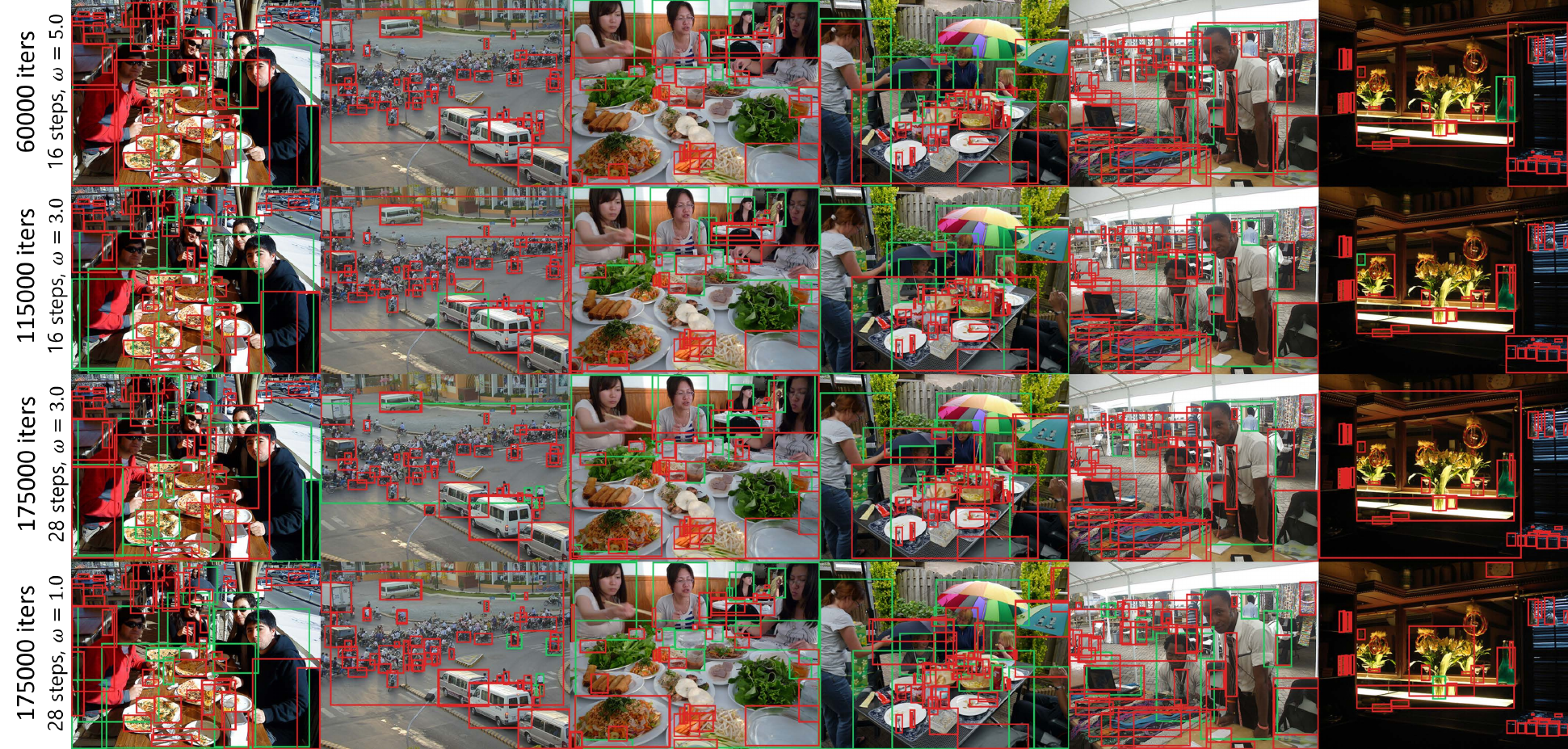}
    \caption{%
    Qualitative YOLOv8m predictions on six COCO test frames, with detections evaluated against ground truth at IoU $\ge 0.5$ and class-aware matching:
    \textcolor[RGB]{34,197,94}{\textbf{green}} boxes are true positives, and \textcolor[RGB]{220,38,38}{\textbf{red}} boxes mark errors of either kind, a predicted box with no matching ground-truth object (false positive) \emph{or} a ground-truth object that no prediction recovered (false negative).
    Each row is a detector trained on a different synthetic pool of \Cref{tab:syn-map}, ordered by ascending test mAP.
    Reading down the rows, red boxes thin out and green ones fill in, which makes one end-to-end fact visible: \emph{which} synthetic pool the detector is trained on determines downstream detection quality, despite all pools using the same generator, the same number of synthetic images ($10{,}000$), and the same locked YOLOv8m recipe.%
    }
    \vspace{-6mm}
    \label{fig:test-pred-grid-coco}
\end{figure*}

\vspace{-1mm}
\subsection{Problem setup}
\label{sec:method:setup}
\vspace{-1mm}
Let $\mathcal{C}$ be a candidate synthetic training set and $\mathcal{T}$ a reference set of real images from the deployment domain. Each image $x$ carries a pretrained feature embedding $\phi(x) \in \mathbb{R}^d$ and per-image metadata $\mu(x)$ read from its bounding-box annotations: object count, box scale (the mean bounding-box-area ratio), and dominant-class label. We seek a score $f(\mathcal{C}, \mathcal{T})$ that requires no detector training and ranks candidate sets in the same order as the downstream mAP of a detector trained on each of them.

\subsection{Conditional-Composition Domain Match}
\label{sec:method:ccdm}
\vspace{-2mm}
\smallbreak
\noindent
\textbf{Grouping images into strata.}
Partition $\mathcal{C}$ and $\mathcal{T}$ into strata $\{S_k\}_{k=1}^{K}$, groups of images whose metadata falls in the same bins. 
Our default stratification, \emph{shared}, uses object-count bins (\textit{solo}: $\leq 1$, \textit{few}: $2$--$5$, \textit{crowded}: $\geq 6$) crossed with box-scale bins ($< 0.001$, $< 0.005$, $< 0.02$, and $\geq 0.02$ of image area), yielding up to $K = 12$ strata.
A second stratification, \emph{class-aware}, refines this by also splitting each stratum by dominant class. This resolves composition more finely, at the cost of fewer images per stratum.
\Cref{fig:bias-axes} shows the three metadata axes on real reference images.

\smallbreak
\noindent
\textbf{The CCDM score.}
Let $\mathcal{C}_k, \mathcal{T}_k$ denote the subsets of $\mathcal{C}, \mathcal{T}$ in stratum $k$, $p_k^{\mathcal{C}} = |\mathcal{C}_k|/|\mathcal{C}|$ and $p_k^{\mathcal{T}} = |\mathcal{T}_k|/|\mathcal{T}|$ the stratum proportions, the fraction of each set's images that lands in stratum $k$, and $D_{\text{base}}$ any feature-space distance from \Cref{sec:preliminary} (\eg, FID). 
CCDM combines a within-stratum appearance term with a composition-mismatch term, 
\begin{equation}
\mathrm{CCDM}(\mathcal{C}, \mathcal{T})
\;=\; D_{\text{cond}} \;+\; D_{\text{global}} \cdot D_{\text{comp}},
\label{eq:ccdm}
\end{equation}
\vspace{-1mm}
where
\begin{align*}
D_{\text{cond}}
&= \sum_{k \in \mathcal{K}^{\star}}
   \tfrac{p_k^{\mathcal{T}}}{\sum_{j \in \mathcal{K}^{\star}} p_j^{\mathcal{T}}}
   \; D_{\text{base}}\!\big(\phi(\mathcal{C}_k), \phi(\mathcal{T}_k)\big),\\
D_{\text{global}}
&= D_{\text{base}}\!\big(\phi(\mathcal{C}), \phi(\mathcal{T})\big),\\
D_{\text{comp}}
&= \mathrm{JSD}\!\big(p^{\mathcal{C}} \,\|\, p^{\mathcal{T}}\big),
\end{align*}
and $\mathcal{K}^{\star}$ is the set of strata with at least four images in both $\mathcal{C}_k$ and $\mathcal{T}_k$ (we set $D_{\text{cond}} = D_{\text{global}}$ when $\mathcal{K}^{\star}$ is empty). 
The first term measures appearance match only within composition-matched strata. The second scales the composition mismatch, the Jensen-Shannon divergence (JSD) between the two sets' stratum proportions, by the global appearance distance, so both terms carry the units of the base distance and can be added whichever base metric is used.


\smallbreak
\noindent
\textbf{CCDM widens the margin on the bias test.}
The margin is the mean-rank gap between the cross-domain matched and in-domain
matched groups, so a larger positive value means the metric places the in-domain group
further ahead.
Re-running the bias test of \Cref{sec:preliminary} with the CCDM variants widens this margin
for every balanced variant, from $0.25$ ranks for the closest classical metric in
that test (GW\textsubscript{DINO}) to $3.50$ for the CCDM version of GW\textsubscript{CLIP},
a base metric the same test calls distribution-biased.
\Cref{tab:ccdm-margin} reports all four.

\begin{table}[t]
\centering
\caption{Mean rank on the bias test of \Cref{sec:preliminary} for the classical metric closest to a correct ordering (top row) and the three balanced CCDM variants. Rank $1$ is the candidate the metric judges closest to the reference, so a good metric gives the two matched candidates low ranks and the collapsed candidate a high one. Applying CCDM to GW\textsubscript{CLIP} raises its own margin from $-5.25$ (\Cref{tab:metric-rankings}) to $+3.50$, well above the $+0.25$ of GW\textsubscript{DINO}, the only balanced classical metric.}
\vspace{-2mm}
\label{tab:ccdm-margin}
\setlength{\tabcolsep}{4pt}
\resizebox{\columnwidth}{!}{%
\begin{tabular}{@{}lcccc@{}}
\toprule
Metric &
\shortstack{In-dom.\\matched} &
\shortstack{Cross-dom.\\matched} &
\shortstack{In-dom.\\collapsed} &
Margin \\
\midrule
GW\textsubscript{DINO} (classical) & 5.00 & 5.25 & 13.25 & $+0.25$ \\
\midrule
$\mathrm{CCDM\text{-}GW}_{\mathrm{CLIP}}$                       & 3.00 & 6.50 &  8.50 & $\mathbf{+3.50}$ \\
$\mathrm{CCDM\text{-}MMD\text{-}Multiscale}_{\mathrm{CLIP}}$     & 3.25 & 4.00 & 10.50 & $+0.75$ \\
$\mathrm{CCDM\text{-}MMD}_{\mathrm{CLIP}}$                 & 3.50 & 4.00 & 10.25 & $+0.50$ \\
\bottomrule
\vspace{-13mm}
\end{tabular}%
}
\end{table}


\vspace{-3mm}
\section{Experiments}
\label{sec:experiments}

\vspace{-1mm}
\subsection{Setup}
\label{sec:exp:setup}
\vspace{-1mm}
We evaluate on COCO~\cite{lin2014microsoft} and VisDrone-DET~\cite{du2019visdrone}. Since COCO's official test-dev annotations are not public, we use \texttt{val2017} ($5{,}000$ images) as the test set. Each synthetic pool is generated from $10{,}000$ real \texttt{train2017} images, class-stratified to preserve the \texttt{train2017} class distribution, rather than from the full ${\sim}118{,}000$-image split. This low-data regime, comparable in scale to VisDrone-DET's $6{,}471$-image train split, sharpens the discrimination test, as differences in synthetic quality translate more directly into downstream performance.
A separate set of $5{,}000$ real images, drawn from the part of \texttt{train2017} not used for generation, serves as the CCDM reference set $\mathcal{T}$. It therefore shares no image with the real images that conditioned the synthetic pools, and it is class-stratified the same way so that it matches them in class distribution. For VisDrone-DET, we use the standard $6{,}471$ train / $548$ val / $1{,}610$ test-dev splits.
Synthetic data comes from a FLUX.1-dev~\cite{flux2024} model finetuned on the COCO and VisDrone-DET train set via LoRA~\cite{hu2022lora} adapters.
Each synthetic image is generated under a condition derived from a real training image, so the pools within a dataset share a common layout structure and differ only in visual appearance. This places the main benchmark in the \emph{fixed-composition regime} of \Cref{sec:intro}: composition is held essentially constant across pools, so a useful metric must resolve appearance differences within matched composition rather than detect a composition shift. For each dataset, we render 18 generator pools by crossing three LoRA checkpoint iterations, three classifier-free guidance scales $\omega \in \{1, 3, 5\}$, which set how strongly the sampler follows the prompt, and two sampling-step counts, $16$ and $28$. The supplementary shows example generations. Guidance scales and step counts are identical across the two datasets. Only the checkpoint iterations differ, $60{,}000$, $115{,}000$, and $175{,}000$ for COCO against $1{,}000$, $3{,}000$, and $5{,}000$ for VisDrone-DET.
Each COCO pool contains $10{,}000$ synthetic images and each VisDrone pool $6{,}471$.
As real-data reference points, we additionally include real candidates. On COCO, we include two, a YOLOv8m retrained from scratch on the $10{,}000$ real images matched to the synthetic pools in index and count, and the official full-\texttt{train2017} YOLOv8m as an upper reference. On VisDrone-DET, we include one, trained from scratch on the real train split. This gives $20$ candidates on COCO ($18$ synthetic pools $+\,2$ real) and $19$ on VisDrone-DET ($18$ $+\,1$). All candidates except the full-\texttt{train2017} reference are trained from scratch under the same locked recipe, so their mAP differences reflect only the training data.
For each candidate, we train YOLOv8m~\cite{jocher2023ultralytics} from scratch, so that the downstream mAP reflects the quality of the training data rather than pretrained priors. The recipe is locked across all runs within a dataset, and only the training set varies. For COCO, we use image size $640$, batch $192$, SGD with a cosine learning-rate schedule (initial lr $0.01$), $5$ warmup epochs, and early stopping with patience $10$ (up to $300$ epochs) on a validation split that we hold out from \texttt{train2017}, since \texttt{val2017} is our test set. We report mAP@$0.5{:}0.95$ on COCO \texttt{val2017} and VisDrone test-dev as our main evaluation metric.
For VisDrone, the val split ($548$ images) serves both as the early-stopping split and as the reference set $\mathcal{T}$, and is disjoint from the test set.

\begin{table*}[t]
\vspace{-2mm}
\centering
\caption{Signed Spearman $\rho$ between each training-free metric and downstream mAP (COCO \texttt{val2017}, VisDrone-DET test-dev), over $20$ candidates on COCO ($18$ synthetic pools $+\,2$ real) and $19$ on VisDrone-DET ($18$ $+\,1$ real). Because every metric here is a distance or a divergence, we negate Spearman $\rho$ for all of them, so a larger $\rho$ always means agreement with higher mAP (\Cref{sec:prelim:polarity}). Metrics are grouped by the three bias categories of \Cref{sec:preliminary}. The best variant per dataset is in bold.}
\vspace{-3mm}
\label{tab:syn-ranking}
\setlength{\tabcolsep}{5pt}
\small
\begin{tabular}{@{}lccc|ccc@{}}
  \toprule
   & \multicolumn{3}{c|}{COCO} & \multicolumn{3}{c}{VisDrone-DET} \\
  \cmidrule(lr){2-4}\cmidrule(lr){5-7}
  Metric & Classical & CCDM-shared & CCDM-classaware & Classical & CCDM-shared & CCDM-classaware \\
  \midrule
  \multicolumn{7}{@{}l}{\emph{appearance-biased}} \\
  \midrule
  FID                               & $+0.912$ & $+0.798$ & $+0.007$ & $+0.889$ & $+0.895$ & $+0.886$ \\
  KID                               & $+0.888$ & $+0.935$ & $+0.938$ & $+0.833$ & $+0.856$ & $+0.862$ \\
  FCD$_\textrm{CLIP}$               & $+0.947$ & $+0.951$ & $+0.915$ & $+0.814$ & $+0.781$ & $+0.781$ \\
  DINO-FD                           & $+0.666$ & $+0.633$ & $+0.317$ & $+0.880$ & $+0.870$ & $+0.851$ \\
  SWD$_\textrm{CLIP}$               & $+0.951$ & $+0.944$ & $+0.941$ & $+0.682$ & $+0.730$ & $+0.713$ \\
  SWD$_\textrm{DINO}$               & $+0.463$ & $+0.581$ & $+0.627$ & $+0.928$ & $+0.934$ & $+0.928$ \\
  CMD$_\textrm{CLIP}$               & $+0.905$ & $+0.933$ & $+0.947$ & $+0.779$ & $+0.800$ & $+0.822$ \\
  CMD$_\textrm{DINO}$               & $+0.613$ & $+0.622$ & $+0.685$ & $+0.911$ & $+0.932$ & $+0.932$ \\
  Wasserstein$_\textrm{CLIP}$       & $+0.201$ & $+0.323$ & $+0.436$ & $+0.618$ & $+0.653$ & $+0.558$ \\
  Wasserstein$_\textrm{DINO}$       & $-0.392$ & $-0.411$ & $-0.171$ & $+0.876$ & $+0.833$ & $+0.767$ \\
  \midrule
  \multicolumn{7}{@{}l}{\emph{distribution-biased}} \\
  \midrule
  MMD-RBF$_\textrm{CLIP}$           & $+0.921$ & $+0.866$ & $+0.970$ & $+0.853$ & $+0.911$ & $+0.862$ \\
  MMD-RBF$_\textrm{DINO}$           & $+0.563$ & $+0.529$ & $+0.828$ & $+0.946$ & $\mathbf{+0.957}$ & $+0.909$ \\
  MMD-RBF-multiscale$_\textrm{CLIP}$& $+0.918$ & $+0.866$ & $\mathbf{+0.976}$ & $+0.862$ & $+0.911$ & $+0.862$ \\
  MMD-RBF-multiscale$_\textrm{DINO}$& $+0.584$ & $+0.532$ & $+0.848$ & $+0.942$ & $+0.948$ & $+0.920$ \\
  GW$_\textrm{CLIP}$                & $+0.362$ & $+0.866$ & $+0.762$ & $-0.478$ & $+0.174$ & $+0.294$ \\
  \midrule
  \multicolumn{7}{@{}l}{\emph{balanced}} \\
  \midrule
  GW$_\textrm{DINO}$                & $+0.759$ & $+0.769$ & $+0.738$ & $-0.375$ & $-0.393$ & $-0.418$ \\
  \bottomrule
  \end{tabular}
\vspace{-4mm}
\end{table*}

\begin{table}[t]
\centering
\renewcommand{\arraystretch}{0.6}
\setlength{\tabcolsep}{4pt}
\caption{mAP@$0.5{:}0.95$ of YOLOv8m trained on each candidate training set, $20$ on COCO ($18$ synthetic pools $+\,2$ real) and $19$ on VisDrone-DET ($18$ $+\,1$), reported on COCO \texttt{val2017} and VisDrone-DET test-dev, alongside the best CCDM variant per dataset (``CCDM''). Pools cross checkpoint levels $C_1{<}C_2{<}C_3$, guidance $\omega$, and sampling steps. CCDM is the class-aware MMD-RBF-multiscale on CLIP for COCO and the shared MMD-RBF on DINO for VisDrone-DET. Lower CCDM is closer to the reference. Both real COCO candidates are scored on the same $10{,}000$ real \texttt{train2017} images and so share one CCDM value.}
\vspace{-2mm}
\label{tab:syn-map}
\resizebox{1.0\columnwidth}{!}{%
\begin{tabular}{@{}>{\scriptsize}c>{\scriptsize}c>{\scriptsize}c|>{\scriptsize}c>{\scriptsize}c|>{\scriptsize}c>{\scriptsize}c@{}}
  \toprule
  \multicolumn{3}{c|}{\normalsize Candidate} & \multicolumn{2}{c|}{\normalsize COCO} & \multicolumn{2}{c}{\normalsize VisDrone} \\
  \cmidrule(lr){1-3}\cmidrule(lr){4-5}\cmidrule(lr){6-7}
  {\normalsize iters} & {\normalsize $\omega$} & {\normalsize steps} & {\normalsize mAP} & {\normalsize best CCDM} & {\normalsize mAP} & {\normalsize best CCDM} \\
  \midrule
  \multicolumn{3}{@{}l|}{Real (full train)} & $0.500$ & $0.0077$ & $0.271$ & $0.0316$ \\
  \multicolumn{3}{@{}l|}{Real ($10$k, scratch)} & $0.246$ & $0.0077$ & -- & -- \\
  \midrule
  $C_1$ & $1$ & $16$ & $0.183$ & $0.0151$ & $0.177$ & $0.0398$ \\
  $C_1$ & $1$ & $28$ & $0.176$ & $0.0138$ & $0.170$ & $0.0363$ \\
  $C_1$ & $3$ & $16$ & $0.155$ & $0.0198$ & $0.131$ & $0.0529$ \\
  $C_1$ & $3$ & $28$ & $0.159$ & $0.0190$ & $0.121$ & $0.0515$ \\
  $C_1$ & $5$ & $16$ & $0.111$ & $0.0281$ & $0.112$ & $0.0713$ \\
  $C_1$ & $5$ & $28$ & $0.121$ & $0.0276$ & $0.112$ & $0.0711$ \\
  \midrule
  $C_2$ & $1$ & $16$ & $0.188$ & $0.0158$ & $0.184$ & $0.0338$ \\
  $C_2$ & $1$ & $28$ & $0.199$ & $0.0127$ & $0.187$ & $0.0303$ \\
  $C_2$ & $3$ & $16$ & $0.164$ & $0.0169$ & $0.126$ & $0.0486$ \\
  $C_2$ & $3$ & $28$ & $0.156$ & $0.0182$ & $0.126$ & $0.0455$ \\
  $C_2$ & $5$ & $16$ & $0.123$ & $0.0254$ & $0.121$ & $0.0619$ \\
  $C_2$ & $5$ & $28$ & $0.124$ & $0.0246$ & $0.117$ & $0.0614$ \\
  \midrule
  $C_3$ & $1$ & $16$ & $0.188$ & $0.0134$ & $0.180$ & $0.0363$ \\
  $C_3$ & $1$ & $28$ & $0.194$ & $0.0124$ & $0.180$ & $0.0340$ \\
  $C_3$ & $3$ & $16$ & $0.162$ & $0.0175$ & $0.141$ & $0.0482$ \\
  $C_3$ & $3$ & $28$ & $0.168$ & $0.0178$ & $0.133$ & $0.0448$ \\
  $C_3$ & $5$ & $16$ & $0.110$ & $0.0255$ & $0.116$ & $0.0645$ \\
  $C_3$ & $5$ & $28$ & $0.117$ & $0.0248$ & $0.124$ & $0.0645$ \\
  \bottomrule
  \end{tabular}
}
\vspace{-5mm}

\end{table}

\vspace{-1mm}
\subsection{Ranking candidate training sets}
\label{sec:exp:syn-ranking}
\Cref{tab:syn-map} reports, for each candidate, the downstream mAP and the best CCDM variant per dataset (class-aware MMD-RBF-multiscale on CLIP for COCO, shared MMD-RBF on DINO for VisDrone-DET), illustrating that CCDM tracks mAP candidate by candidate. On VisDrone-DET, the real candidate attains the highest mAP and one of the two lowest CCDM scores ($0.032$), outperforming every synthetic pool by at least $0.08$ mAP, while the low-guidance ($\omega = 1$) pools beat the $\omega = 3$ pools by up to $0.06$ mAP. Ordering the candidates by these CCDM variants agrees with the mAP ordering at $\rho = +0.976$ on COCO and $+0.957$ on VisDrone-DET. \Cref{fig:test-pred-grid-coco} shows this qualitatively: true-positive (green) boxes increase and error (red) boxes decrease as the pool's mAP rises.

\Cref{tab:syn-ranking} extends the comparison to all classical metrics and their CCDM variants, reporting signed Spearman $\rho$ over the candidate mAP ordering on both datasets ($20$ candidates on COCO, $19$ on VisDrone-DET). Both benchmarks hold composition fixed, so they test only whether a metric can separate candidates that differ in appearance alone. Sensitivity to composition shift is evaluated separately in \Cref{tab:ccdm-margin}. Four patterns emerge.
\smallbreak
\noindent
\textbf{Class-aware CCDM ranks candidates best.}
Class-aware CCDM on MMD-RBF-multiscale$_\textrm{CLIP}$ reaches $\rho = +0.976$ over the 20 candidates, the highest of all metric variants.
Class-aware conditioning sharpens the distribution-biased metrics the most: MMD-RBF$_\textrm{CLIP}$ rises from $+0.921$ to $+0.970$, MMD-RBF$_\textrm{DINO}$ from $+0.563$ to $+0.828$, and GW$_\textrm{CLIP}$ from $+0.362$ to $+0.762$, confirming that stratifying by composition before measuring appearance tightens the correlation with mAP.

\smallbreak
\noindent
\textbf{Classical appearance metrics are also strong on COCO.}
On COCO the benchmark sits squarely in the fixed-composition regime, so even a raw global appearance distance already aligns with pool quality (SWD$_\textrm{CLIP}$ $+0.951$, FCD$_\textrm{CLIP}$ $+0.947$). The question here is therefore not whether a metric beats chance but whether conditioning still helps. It does: the best CCDM variant surpasses the best classical metric ($+0.976$ vs.\ $+0.951$), and the gap widens for the distribution-biased metrics whose raw scores are weaker.
\Cref{fig:metric-vs-map-coco} illustrates this against FID~\cite{heusel2017gans}.
\smallbreak
\noindent
\textbf{Class-aware outperforms shared at COCO's reference size.}
With $5{,}000$ reference images, the finer count$\times$scale$\times$class strata stay above the four-image floor required by $D_{\text{cond}}$ (\Cref{sec:method:ccdm}), so splitting each stratum further by dominant class raises MMD-RBF$_\textrm{CLIP}$ from $+0.866$ to $+0.970$ and MMD-RBF-multiscale$_\textrm{CLIP}$ from $+0.866$ to $+0.976$.
This reverses the VisDrone-DET trend, where only $548$ val images leave most fine strata under-populated and the coarser CCDM-shared is more reliable.
Stratification granularity should scale with reference size.
\smallbreak
\noindent
\textbf{Class-aware conditioning favors kernel metrics over moment matching.}
Class-aware strata help kernel-based metrics (KID, MMD) but break moment-matching ones: FID collapses from $+0.912$ to $+0.007$ and DINO-FD from $+0.666$ to $+0.317$, as the Gaussian moment estimate degrades on few images per stratum, while kernel-based metrics remain robust.
This motivates kernel metrics as the CCDM base.
The leading CCDM variant differs by dataset, class-aware MMD-RBF-multiscale$_\textrm{CLIP}$ on COCO and shared MMD-RBF$_\textrm{DINO}$ on VisDrone-DET, but on each dataset a CCDM variant surpasses every classical baseline. 
A metadata-only score cannot be compared here: all 18 pools share the
same COCO layouts, so the class-distribution distance is identical across them
($0.0372$) and no ranking exists. It becomes informative only once composition
varies (\Cref{tab:supp-compshift}).

\begin{figure}[t]
    \centering
    \includegraphics[width=\columnwidth]{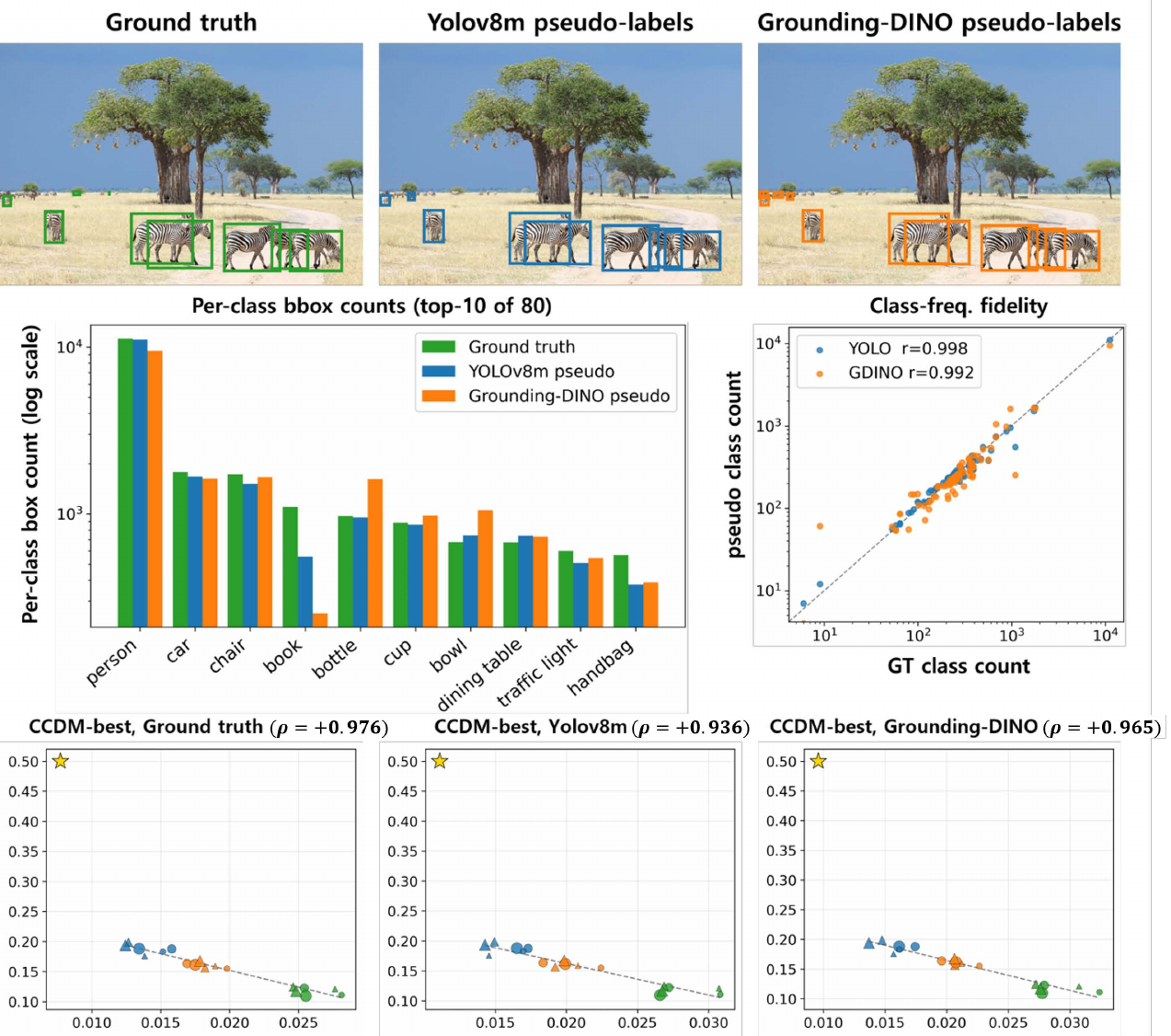}
    \caption{\textbf{CCDM with an unlabeled reference set.} Best CCDM $=$ class-aware MMD-RBF-multiscale$_\textrm{CLIP}$. \emph{Top:} a real reference image (COCO \texttt{train2017}) with ground-truth, YOLOv8m, and Grounding-DINO boxes. \emph{Middle:} per-class counts (top-$10$, log) and class-frequency fidelity (Pearson $r=0.998$/$0.992$ for YOLO/G-DINO). \emph{Bottom:} best CCDM vs.\ mAP per metadata source ($\rho=+0.976$/$+0.936$/$+0.965$). Even with pseudo-labels it exceeds FID ($\rho=+0.912$), so the ranking survives.}
    \vspace{-3mm}
    \label{fig:annot-free}
\end{figure}

\subsection{CCDM with an unlabeled reference set}
\vspace{-1mm}
\label{sec:exp:annot-free}
We now assume access to real reference images but no ground-truth annotations, since obtaining annotations is expensive. As CCDM relies on per-image metadata, we ask whether it remains usable when this metadata must be estimated by an off-the-shelf detector. We obtain pseudo-metadata by running a detector on the unlabeled reference images. To ensure the conclusion does not depend on a particular detector, we use both a closed-set model (YOLOv8m) and an open-vocabulary one (Grounding-DINO). Pseudo-labeling is applied only to the reference set, as the synthetic candidates already carry metadata from their layout conditions.
Such pseudo-metadata is inevitably noisy, as the detectors disagree with both the ground truth and each other on individual boxes and classes. The relevant question is not how accurate the pseudo-labels are, but whether CCDM's ranking of candidate training sets survives this noise. It does. Ground-truth metadata yields $\rho = +0.976$, and switching to detector pseudo-labels barely changes it, $\rho = +0.936$ with YOLOv8m and $+0.965$ with Grounding-DINO (\Cref{tab:annot-free}), a drop of at most $0.040$. This holds because the pseudo-metadata tracks the ground truth at the population level (per-class frequency Pearson $r = 0.998$ and $0.992$ for the two detectors, \Cref{fig:annot-free}), even though individual boxes are imperfect. CCDM therefore does not need precise metadata, only metadata that preserves the relative composition across candidates.

\begin{table}[t]
\centering
\small
\caption{CCDM ranking ability (signed Spearman $\rho$ with COCO \texttt{val2017} mAP) under ground-truth vs.\ detector pseudo-label metadata.
\textbf{Src}: metadata source. GT (ground truth), YOLO (YOLOv8m pseudo-labels), G-DINO (Grounding-DINO pseudo-labels).
\textbf{CCDM$_{\mathrm{CLIP}}$}/\textbf{CCDM$_{\mathrm{DINO}}$}: class-aware CCDM-MMD-RBF-multiscale on CLIP / DINOv2 features.
\textbf{$r$}: class-frequency Pearson correlation vs.\ GT.
Pseudo-labels retain CCDM's ranking without any ground-truth annotation.}
\vspace{-2mm}
\label{tab:annot-free}
\setlength{\tabcolsep}{8pt}
\begin{tabular}{@{}lccc@{}}
\toprule
Src & CCDM$_{\mathrm{CLIP}}$ & CCDM$_{\mathrm{DINO}}$ & $r$ \\
\midrule
GT     & $+0.976$ & $+0.848$ & -- \\
YOLO   & $+0.936$ & $+0.736$ & $0.998$ \\
G-DINO & $+0.965$ & $+0.876$ & $0.992$ \\
\bottomrule
\end{tabular}
\vspace{-3mm}
\end{table}

\begin{table}[t]
\centering
\caption{Wall-clock cost of assessing synthetic-data quality on a single H200 GPU. Training a YOLOv8m detector from scratch measures quality directly but is expensive, whereas CCDM ranks the same candidates training-free in a fraction of the time. Train-from-scratch time is the per-candidate average, ranging $0.6$--$1.8$\,h depending on the early-stop epoch.}
\vspace{-3mm}
\label{tab:cost}
\resizebox{\columnwidth}{!}{%
\begin{tabular}{@{}lccc@{}}
\toprule
 & Per candidate & All 20 candidates & Speedup \\
\midrule
Train-from-scratch & $1.3$\,h & ${\sim}26$\,h & $1\times$ \\
CCDM (ours)        & $12.7$\,min & $4.2$\,h & $6\times$ \\
\bottomrule
\end{tabular}%
}
\vspace{-6mm}
\end{table}

\subsection{Cost of quality assessment}
\vspace{-1mm}
\label{sec:exp:cost}
Measuring synthetic-data quality by training a detector is accurate but expensive. On a single H200 GPU, training YOLOv8m from scratch takes $1.3$\,h per candidate on average ($0.6$--$1.8$\,h depending on the early-stop epoch, \Cref{tab:cost}), so screening all $20$ candidates costs about $26$\,h. CCDM assesses the same candidates without training, taking only $12.7$\,min per candidate (${\sim}4.2$\,h for all $20$), a $6\times$ speedup. Because CCDM ranks candidates in strong agreement with downstream mAP (\Cref{sec:exp:syn-ranking}) yet costs a fraction of a single training run, it is practical for quickly deciding which synthetic pool to train the final detector on.
\vspace{-2mm}
\section{Conclusion}
\vspace{-1mm}
Classical training-free metrics judge detection training data by one global statistic, which is only weakly sensitive to composition shift and, at fixed composition, averages away the appearance differences that separate strong from weak candidates.
Our Conditional-Composition Domain Match (CCDM) conditions any feature-space distance on per-image metadata, matching appearance within composition-matched strata and adding a composition penalty scaled by the global distance.
On COCO and VisDrone-DET, the best CCDM variant ranks 20 and 19 candidate training sets in strong agreement with YOLOv8 mAP, outperforming FID, KID, and MMD, and it holds when reference metadata comes from off-the-shelf detector pseudo-labels.
Promising directions include further benchmarks (\eg, LVIS~\cite{gupta2019lvis}) and tasks such as instance and panoptic segmentation, where per-image composition is similarly informative.

\clearpage

\noindent
\textbf{Acknowledgement.} 
This research was supported by the MSIT (Ministry of Science and ICT), Korea, under the Defense ICT Innovation Technology program (RS-2026-25539289) supervised by the IITP (Institute of Information \& Communications Technology Planning \& Evaluation).

{
    \small
    \nocite{*}
    \bibliographystyle{ieeenat_fullname}
    \bibliography{main}
}

\clearpage
\clearpage
\setcounter{page}{1}
\maketitlesupplementary
\appendix

\section{Generalization across detector architectures}
\label{sec:supp:detector}

\begin{figure*}[t]
\centering
\begin{subfigure}{\textwidth}
  \centering
  \includegraphics[width=\textwidth]{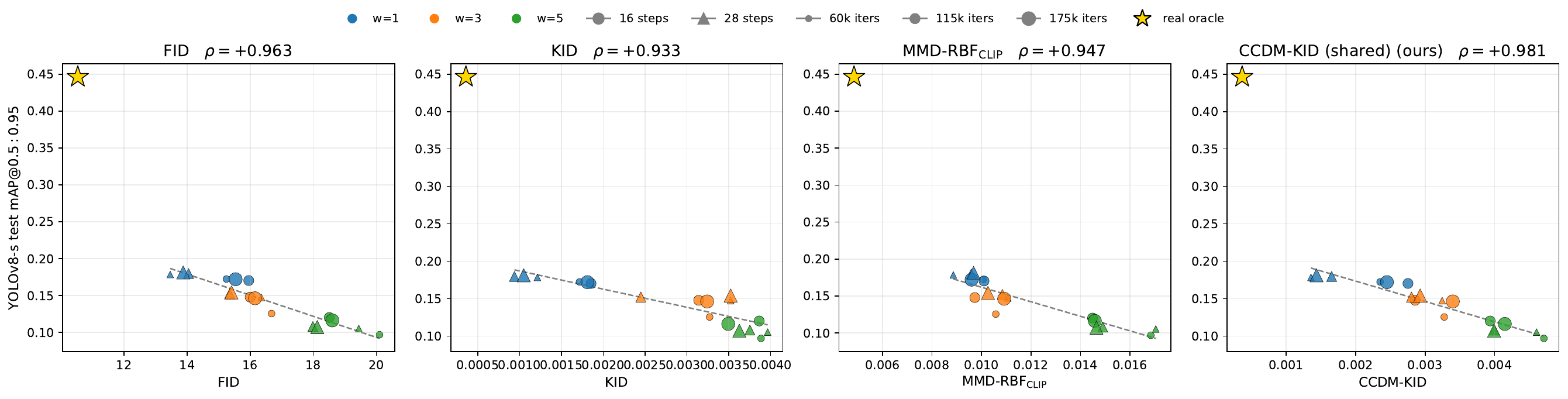}
  \caption{\textbf{YOLOv8-s}. Best-correlating metric is CCDM-KID (shared).}
  \label{fig:scatter-v8s}
\end{subfigure}

\vspace{1.5ex}
\begin{subfigure}{\textwidth}
  \centering
  \includegraphics[width=\textwidth]{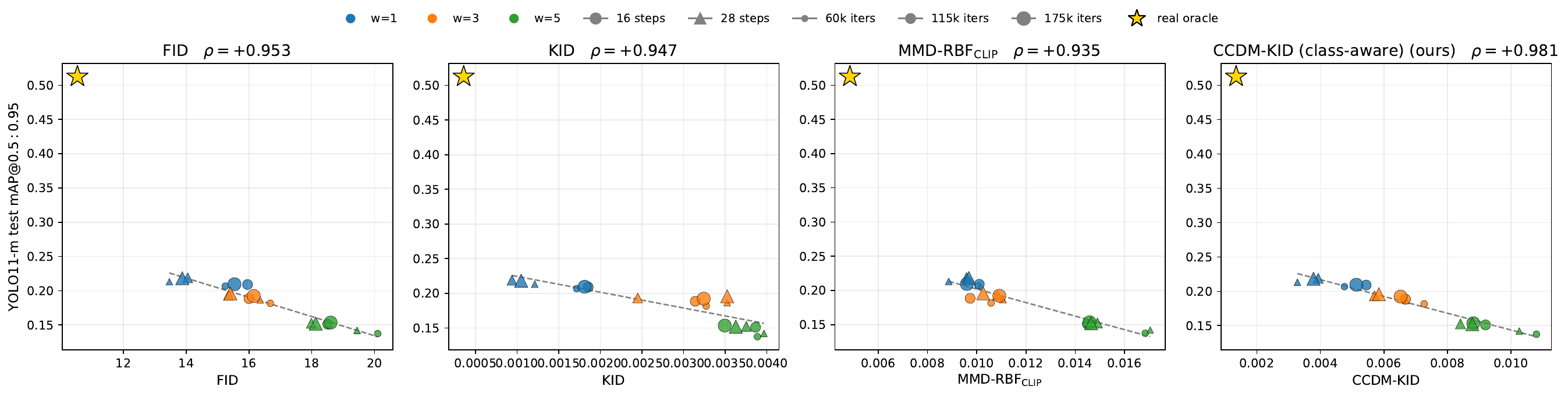}
  \caption{\textbf{YOLO11-m}. Best-correlating metric is CCDM-KID (class-aware).}
  \label{fig:scatter-yolo11}
\end{subfigure}

\vspace{1.5ex}
\begin{subfigure}{\textwidth}
  \centering
  \includegraphics[width=\textwidth]{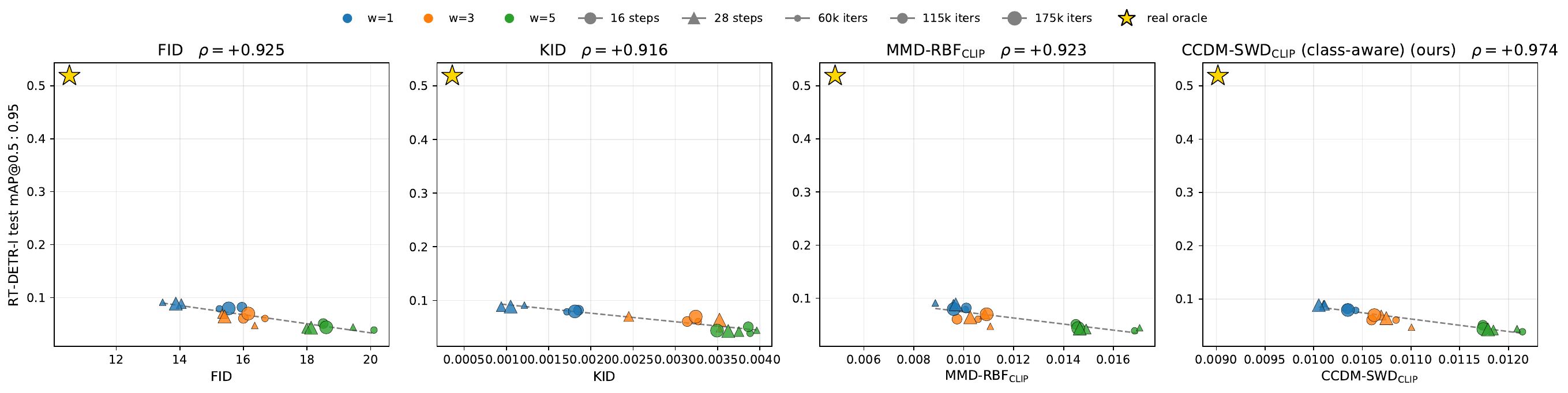}
  \caption{\textbf{RT-DETR-l}. Best-correlating metric is CCDM-SWD on CLIP (class-aware).}
  \label{fig:scatter-rtdetr}
\end{subfigure}
\caption{Per-pool test mAP@$0.5{:}0.95$ vs metric across three detectors on COCO \texttt{val2017} ($18$ synthetic pools plus the real oracle). Each row shows the three widely used classical baselines (FID, KID, MMD-RBF on CLIP) and the best-correlating CCDM variant for that detector. Color encodes guidance $\omega$, marker encodes sampling steps, size encodes checkpoint iters, and the gold star is the real oracle (COCO-pretrained weights). All $\rho$ are computed over the $19$ points.}
\label{fig:scatter-detectors}
\end{figure*}

The main benchmark (\Cref{sec:exp:syn-ranking}) measures downstream quality with a single detector, YOLOv8m. To check that CCDM's advantage is not tied to that detector, we train the same $18$ COCO synthetic pools from scratch with three more detectors, from a smaller model in the same family (YOLOv8-s) to a newer-generation CNN (YOLO11-m) and the transformer-based RT-DETR-l. As in the main benchmark, we add a real oracle, each detector's official COCO-pretrained weights evaluated on val2017 with no synthetic training, as an upper anchor for real-data supervision. The candidate pools, the real reference, and all training-free metrics are held fixed, only the downstream detector varies. \Cref{fig:scatter-v8s,fig:scatter-yolo11,fig:scatter-rtdetr} plot test mAP@$0.5{:}0.95$ against the three representative classical baselines and the best-correlating CCDM variant for each detector, with $\rho$ computed over all $19$ points.

\noindent\textbf{Robustness to detector capacity.} Staying within the YOLOv8 family, shrinking the detector from YOLOv8m to the smaller YOLOv8-s leaves the ranking behavior unchanged. A CCDM variant is still the best predictor of downstream mAP for YOLOv8-s ($\rho=+0.981$), on par with the YOLOv8m benchmark ($\rho=+0.976$), so the metric does not rely on a particular model size.

\noindent\textbf{Generalization across architectures.} Moving beyond YOLOv8, a CCDM variant remains the top-correlating metric for both a newer-generation CNN and a transformer detector, CCDM-KID for YOLO11-m ($\rho=+0.981$) and CCDM-SWD$_\textrm{CLIP}$ for RT-DETR-l ($\rho=+0.974$). The best base distance varies with the detector, which is expected as architectures emphasize different cues. The invariant is therefore not one universal metric, but rather that conditioning through CCDM yields the strongest correlation with downstream mAP in each case.

\noindent\textbf{Conditioning improves most base distances.} CCDM lifts KID from $+0.933$ to $+0.981$ (YOLOv8-s) and from $+0.947$ to $+0.981$ (YOLO11-m), and SWD$_\textrm{CLIP}$ from $+0.951$ to $+0.974$ (RT-DETR-l). Measuring appearance within composition-matched strata, rather than over the pool as a whole, is what lifts the best metric on each detector above the strongest classical baseline there.

\noindent\textbf{Stratification favors kernel metrics.} As in the main benchmark, kernel distances such as KID and MMD gain under class-aware stratification, whereas FID collapses from above $+0.9$ to around zero on every detector, since its Gaussian moment estimate is unreliable on the small per-stratum samples.

\noindent These results indicate that CCDM tracks downstream detection quality independently of detector size, generation, and architecture. Per-metric correlations for all detectors are reported in \Cref{tab:rank-v8s,tab:rank-yolo11,tab:rank-rtdetr}.

\section{Sensitivity to composition shift}
\label{sec:supp:comp-shift}

In the ranking experiments, every synthetic pool shared the same composition and differed only in appearance. Here we probe the opposite case, where appearance is held fixed and only the label composition is corrupted, to show that CCDM responds to the metadata that appearance metrics are blind to. Starting from a fixed pool of $30$K synthetic images, we cumulatively drop the rarest classes from the training labels, fewest-sample first, leaving $80$ down to $10$ classes (\Cref{tab:comp-shift}). The pixels are identical across every row, so only the conditioning metadata changes, while the reference target remains the full real distribution.

As more classes are dropped, the YOLOv8m detector degrades sharply, from mAP $0.325$ at $80$ classes to $0.031$ at $10$, since the labels increasingly omit objects it should learn to find. Every appearance-based classical metric is, by construction, insensitive to this. FID, KID, SWD$_\textrm{CLIP}$, and MMD$_\textrm{CLIP}$ remain numerically identical across all six rows ($11.81$, $0.00109$, $0.00390$, $0.00464$), because they compare only pixels, and the pixels never change. They cannot tell a faithful pool apart from one whose labels have lost most of their classes.

CCDM behaves the opposite way. It reads the composition through its conditioning and rises from $0.0113$ to $0.0158$ as the labels degrade, tracking the drop in mAP with Spearman $\rho=+0.94$. Because CCDM compares appearance within composition-matched strata and carries an explicit composition divergence term, distorting the label distribution directly enlarges the distance to the real reference. This is precisely the property a data-quality metric needs and a pure appearance distance lacks, sensitivity to whether the synthetic metadata still matches the target, not only whether the images look real.

\section{Additional qualitative results}
\label{sec:supp:qualitative}

\begin{figure*}
  \centering
  \includegraphics[width=\textwidth]{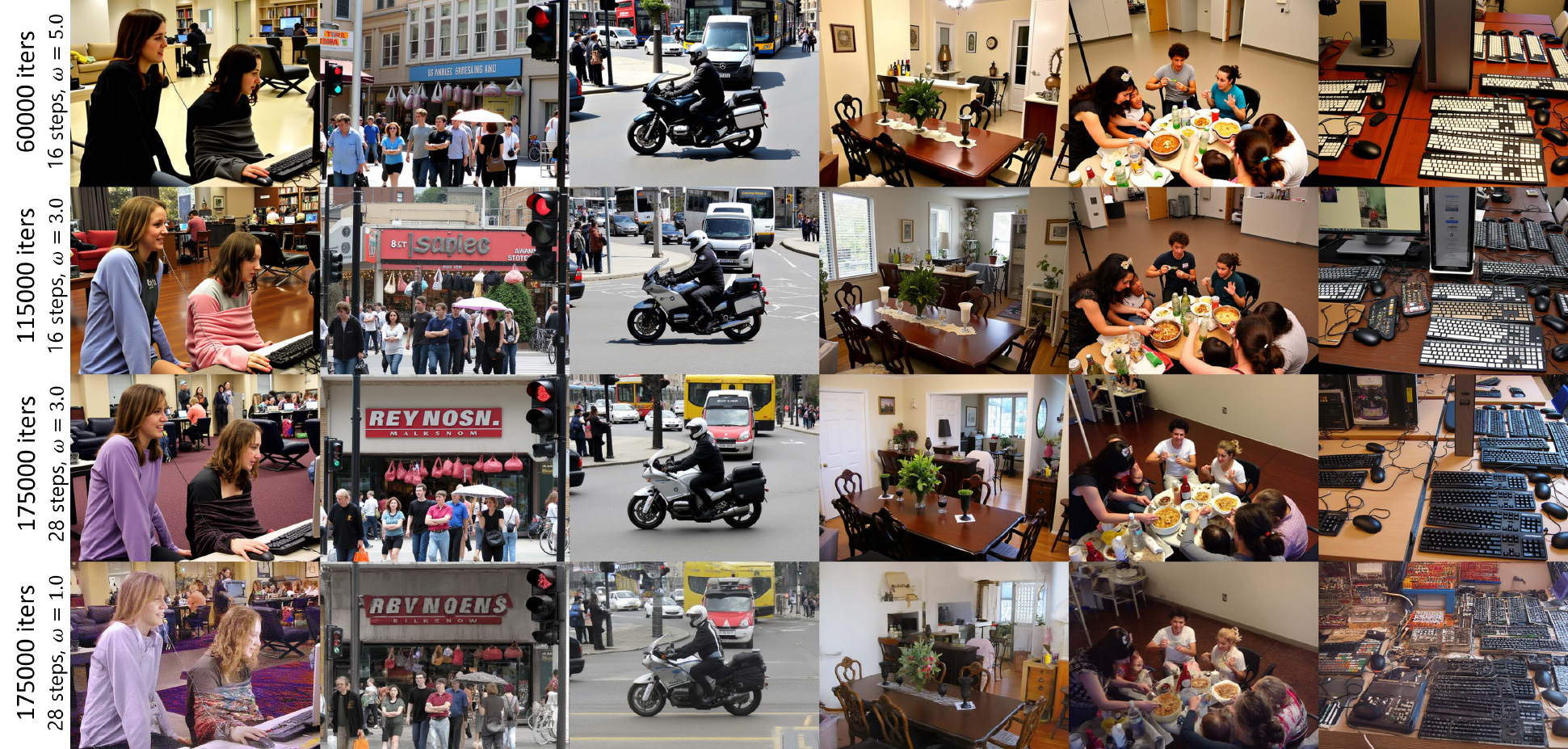}
  \caption{As \Cref{fig:synthetic-comparison}, for the 18-pool COCO sweep. Composition is fixed across pools, yet val2017 mAP@$0.5{:}0.95$ ranges $0.110$ to $0.199$ (\Cref{tab:syn-map}), and a global appearance distance separates them only partially (\Cref{tab:syn-ranking}), motivating composition-matched strata (\Cref{sec:method}).}
  \label{fig:synthetic-comparison-coco}
\end{figure*}

\begin{figure*}
  \centering
  \includegraphics[width=\textwidth]{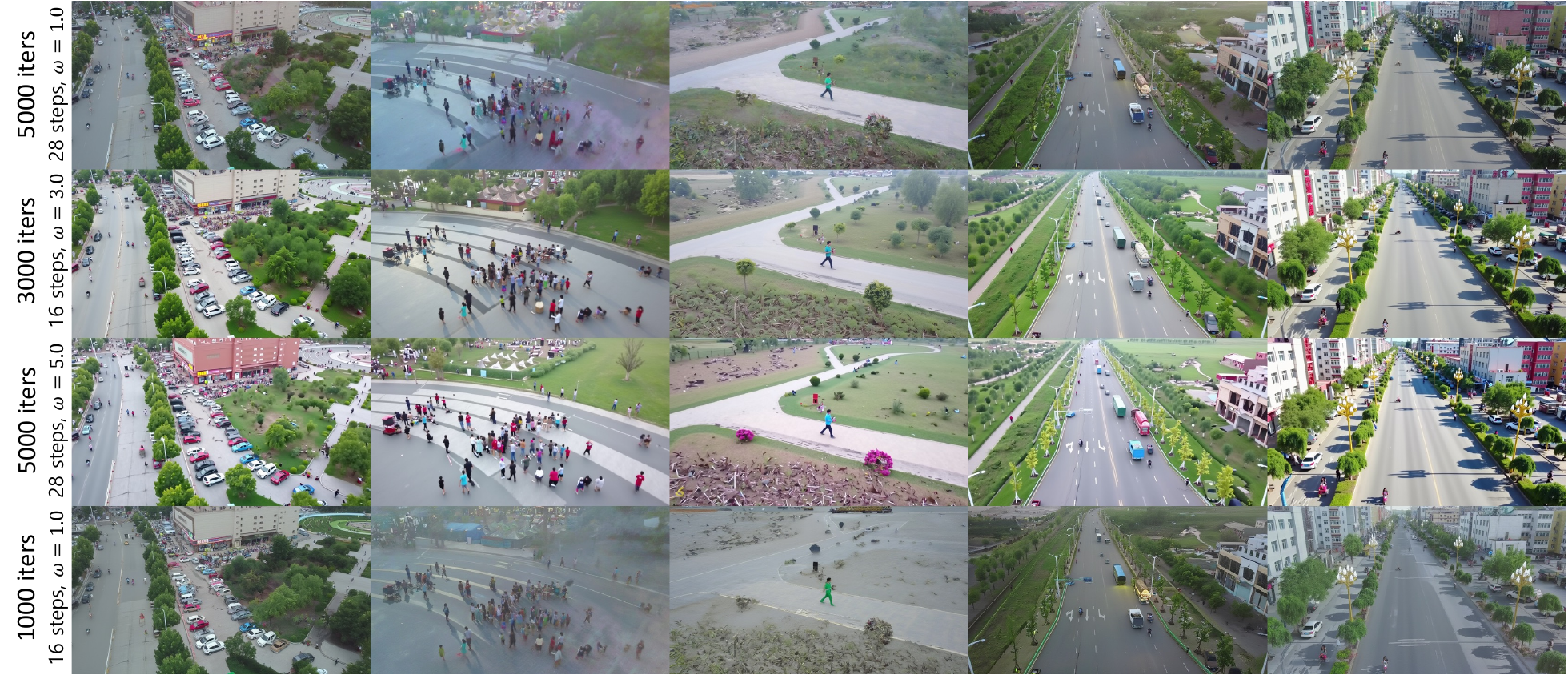}
  \caption{Generations from four representative pools of the 18-pool VisDrone-DET sweep (\Cref{sec:experiments}). Each row is one (checkpoint iteration, guidance $\omega$, sampling steps) setting, with adjacent rows differing on all three axes. Each column is the same conditioning frame, so vertical comparisons isolate appearance. Composition is fixed, yet test-dev mAP varies 67\% relative ($0.112$ to $0.187$, \Cref{tab:syn-map}), and a single global appearance distance separates the pools only partially (\Cref{tab:syn-ranking}), motivating composition-matched strata (\Cref{sec:method}).}
  \label{fig:synthetic-comparison}
\end{figure*}

\begin{figure*}[t]
  \centering
  \includegraphics[width=\textwidth]{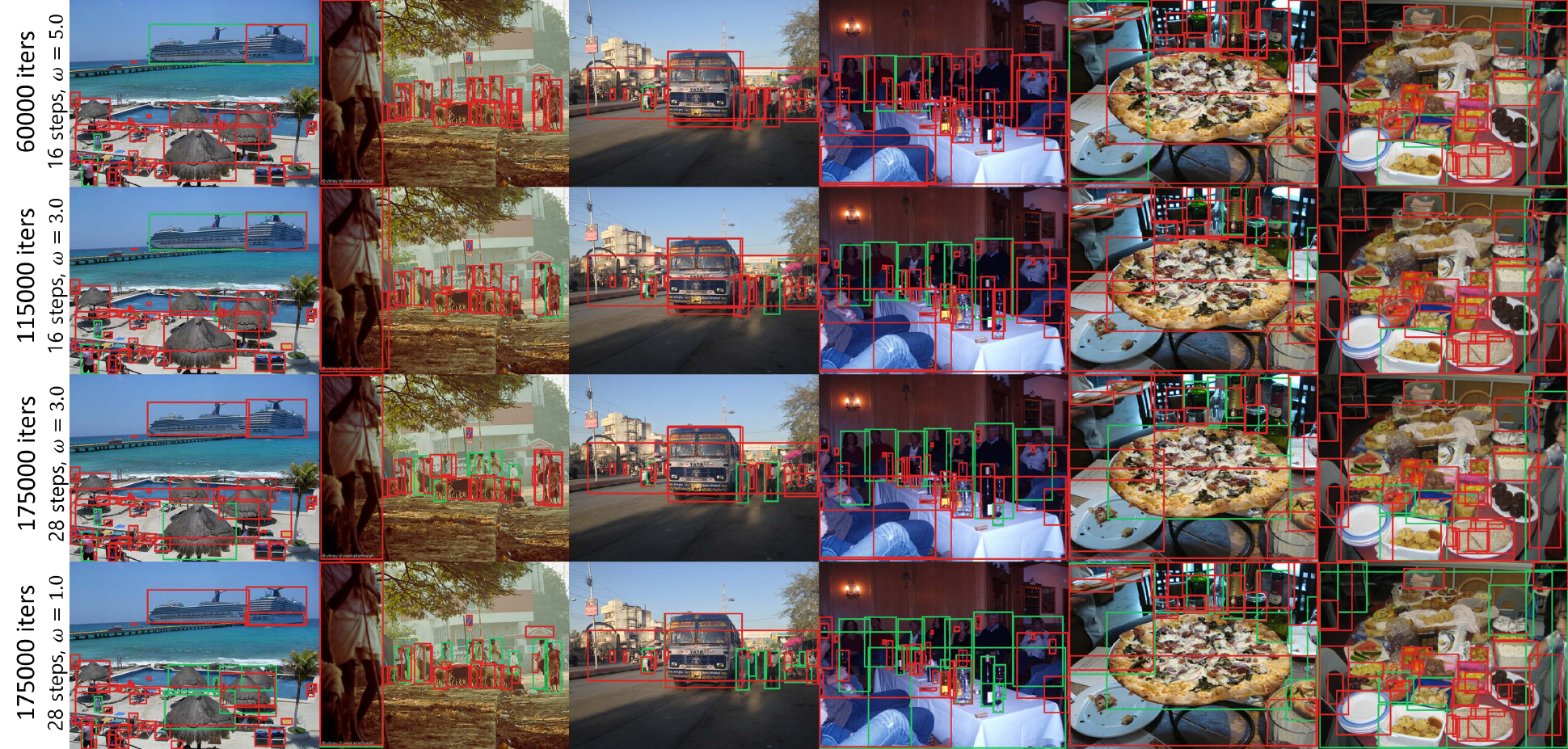}
  \caption{Qualitative YOLOv8m predictions on six COCO \texttt{val2017} frames, evaluated against ground truth at IoU $\ge 0.5$ with class-aware matching. \textcolor[RGB]{34,197,94}{\textbf{Green}} boxes are true positives, and \textcolor[RGB]{220,38,38}{\textbf{red}} boxes mark errors of either kind, a predicted box with no matching ground-truth object (false positive) or a ground-truth object that no prediction recovered (false negative). Rows are four synthetic pools ordered by increasing val2017 mAP@$0.5{:}0.95$ (top $0.111$ to bottom $0.194$), all trained with the same generator and the same locked YOLOv8m recipe. Detections shift from mostly red to mostly green down the rows, even though the pools are hard to tell apart by eye.}
  \label{fig:test-pred-grid-coco-supp}
\end{figure*}

\begin{figure*}[t]
  \centering
  \includegraphics[width=\textwidth]{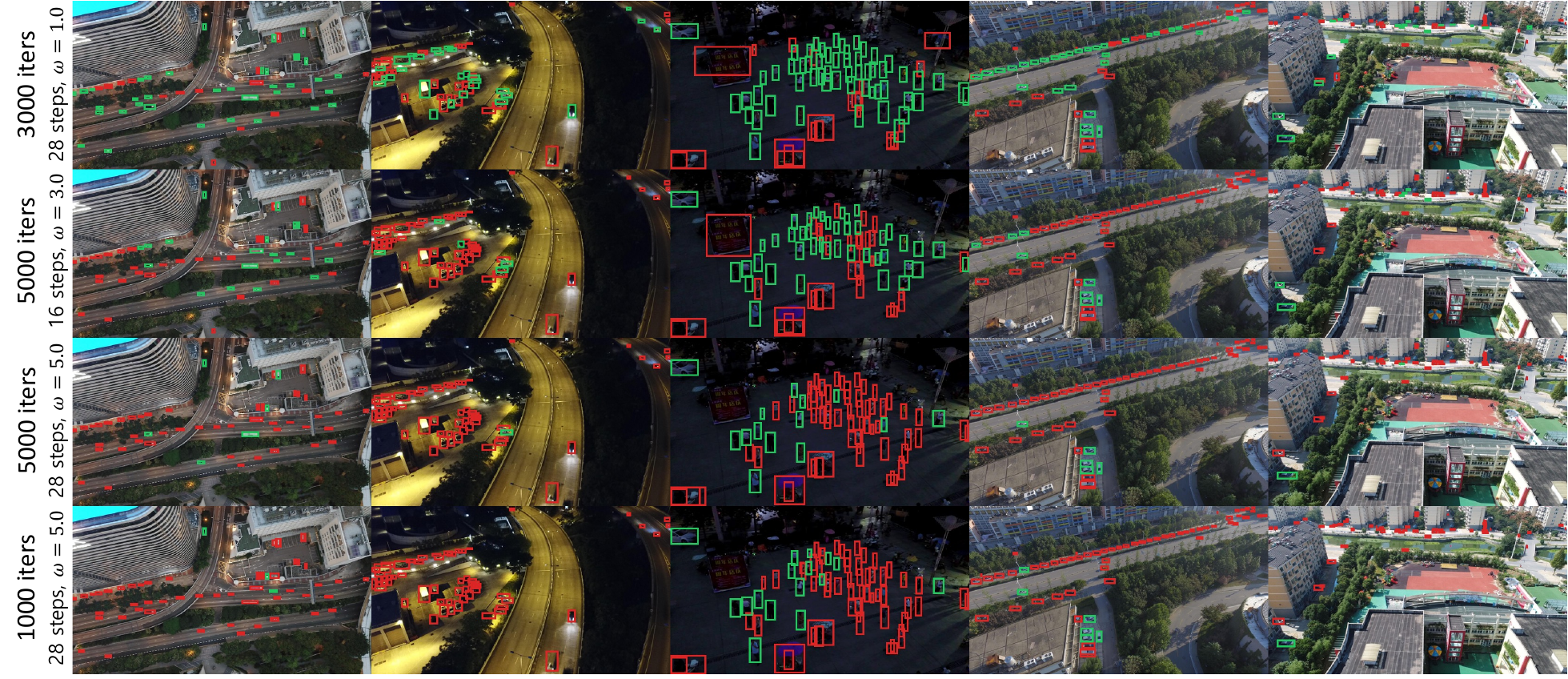}
  \caption{%
    Qualitative YOLOv8m predictions on five VisDrone-DET test-dev frames, with detections evaluated against ground truth at IoU $\ge 0.5$ and class-aware matching:
  \textcolor[RGB]{34,197,94}{\textbf{green}} boxes are true positives, and \textcolor[RGB]{220,38,38}{\textbf{red}} boxes mark errors of either kind, a predicted box with no matching
  ground-truth object (false positive) \emph{or} a ground-truth object that no prediction recovered (false negative). Rows are four synthetic pools ordered by decreasing test-dev mAP@$0.5{:}0.95$ (top $0.187$ to bottom $0.112$). Reading up the rows, red boxes thin out and green ones fill in, which makes one end-to-end fact visible: \emph{which} synthetic pool the detector
  is trained on determines whether downstream detection is competent or near-blank, despite all pools using the same generator, same number of synthetic images ($6{,}471$), and the same
  locked YOLOv8m recipe.
  }
  \label{fig:test-pred-grid-visdrone}
\end{figure*}

\Cref{fig:synthetic-comparison-coco} and \Cref{fig:synthetic-comparison} show example generations from the COCO and VisDrone-DET pools, and \Cref{fig:test-pred-grid-coco-supp} and \Cref{fig:test-pred-grid-visdrone} show detections from YOLOv8m detectors trained on those pools and tested on real images. On both datasets, the generated images look realistic, and the pools are hard to tell apart by eye, so it is not clear from the images which pool trains a better detector. However, the detections differ clearly across pools, even though all pools use the same generator and the same training recipe. This is why a training-free metric is needed, since visual inspection alone does not show which synthetic data is useful for training.

\begin{table}[t]
\centering
\caption{Composition-shift ablation on COCO. From a fixed pool of 30K synthetic images we cumulatively drop the rarest classes (fewest-sample first) from the training labels, leaving the listed number of classes. The images are identical across rows so only the label composition changes, while the reference target stays the full real distribution. As composition degrades, a YOLOv8m detector's mAP@$0.5{:}0.95$ falls and CCDM rises to track it, whereas every appearance-based classical metric stays exactly constant because they only see the unchanged pixels. This isolates composition sensitivity rather than ranking quality, which \Cref{tab:syn-ranking} measures.}
\label{tab:comp-shift}
\setlength{\tabcolsep}{5pt}
\resizebox{\columnwidth}{!}{%
\begin{tabular}{@{}ccccccc@{}}
\toprule
Classes & mAP & CCDM (ours) & FID & KID & SWD$_{\text{CLIP}}$ & MMD$_{\text{CLIP}}$ \\
\midrule
$80$ & $0.325$ & $0.0113$ & $11.81$ & $0.00109$ & $0.00390$ & $0.00464$ \\
$60$ & $0.236$ & $0.0131$ & $11.81$ & $0.00109$ & $0.00390$ & $0.00464$ \\
$40$ & $0.118$ & $0.0141$ & $11.81$ & $0.00109$ & $0.00390$ & $0.00464$ \\
$25$ & $0.073$ & $0.0157$ & $11.81$ & $0.00109$ & $0.00390$ & $0.00464$ \\
$15$ & $0.046$ & $0.0144$ & $11.81$ & $0.00109$ & $0.00390$ & $0.00464$ \\
$10$ & $0.031$ & $0.0158$ & $11.81$ & $0.00109$ & $0.00390$ & $0.00464$ \\
\bottomrule
\end{tabular}%
}
\end{table}

\begin{table}[t]
\centering
\caption{Signed Spearman $\rho$ between each training-free metric and YOLOv8-s downstream mAP@$0.5{:}0.95$ on COCO \texttt{val2017} ($18$ synthetic pools plus the real oracle, $19$ points). Best in bold.}
\label{tab:rank-v8s}
\setlength{\tabcolsep}{5pt}\small
\begin{tabular}{@{}lccc@{}}
\toprule
Metric & Classical & CCDM-shared & CCDM-classaware \\
\midrule
FID & $+0.963$ & $+0.839$ & $-0.030$ \\
KID & $+0.933$ & $\mathbf{+0.981}$ & $+0.972$ \\
FCD$_\textrm{CLIP}$ & $+0.968$ & $+0.972$ & $+0.944$ \\
DINO-FD & $+0.651$ & $+0.605$ & $+0.240$ \\
SWD$_\textrm{CLIP}$ & $+0.965$ & $+0.961$ & $+0.965$ \\
SWD$_\textrm{DINO}$ & $+0.409$ & $+0.560$ & $+0.612$ \\
CMD$_\textrm{CLIP}$ & $+0.914$ & $+0.963$ & $+0.968$ \\
CMD$_\textrm{DINO}$ & $+0.600$ & $+0.595$ & $+0.686$ \\
Wass$_\textrm{CLIP}$ & $+0.132$ & $+0.488$ & $+0.514$ \\
Wass$_\textrm{DINO}$ & $-0.449$ & $-0.396$ & $-0.244$ \\
MMD-RBF$_\textrm{CLIP}$ & $+0.947$ & $+0.898$ & $+0.960$ \\
MMD-RBF$_\textrm{DINO}$ & $+0.591$ & $+0.509$ & $+0.860$ \\
MMD-ms$_\textrm{CLIP}$ & $+0.942$ & $+0.898$ & $+0.961$ \\
MMD-ms$_\textrm{DINO}$ & $+0.616$ & $+0.504$ & $+0.879$ \\
GW$_\textrm{CLIP}$ & $+0.516$ & $+0.854$ & $+0.874$ \\
GW$_\textrm{DINO}$ & $+0.688$ & $+0.747$ & $+0.765$ \\
\bottomrule
\end{tabular}
\end{table}

\begin{table}[t]
\centering
\caption{Signed Spearman $\rho$ between each training-free metric and YOLO11-m downstream mAP@$0.5{:}0.95$ on COCO \texttt{val2017} ($18$ synthetic pools plus the real oracle, $19$ points). Best in bold.}
\label{tab:rank-yolo11}
\setlength{\tabcolsep}{5pt}\small
\begin{tabular}{@{}lccc@{}}
\toprule
Metric & Classical & CCDM-shared & CCDM-classaware \\
\midrule
FID & $+0.953$ & $+0.830$ & $-0.044$ \\
KID & $+0.947$ & $+0.953$ & $\mathbf{+0.981}$ \\
FCD$_\textrm{CLIP}$ & $+0.965$ & $+0.965$ & $+0.939$ \\
DINO-FD & $+0.649$ & $+0.598$ & $+0.225$ \\
SWD$_\textrm{CLIP}$ & $+0.974$ & $+0.972$ & $+0.961$ \\
SWD$_\textrm{DINO}$ & $+0.411$ & $+0.540$ & $+0.595$ \\
CMD$_\textrm{CLIP}$ & $+0.933$ & $+0.967$ & $+0.965$ \\
CMD$_\textrm{DINO}$ & $+0.595$ & $+0.596$ & $+0.674$ \\
Wass$_\textrm{CLIP}$ & $+0.212$ & $+0.498$ & $+0.468$ \\
Wass$_\textrm{DINO}$ & $-0.419$ & $-0.418$ & $-0.242$ \\
MMD-RBF$_\textrm{CLIP}$ & $+0.935$ & $+0.891$ & $+0.974$ \\
MMD-RBF$_\textrm{DINO}$ & $+0.565$ & $+0.504$ & $+0.853$ \\
MMD-ms$_\textrm{CLIP}$ & $+0.930$ & $+0.891$ & $+0.968$ \\
MMD-ms$_\textrm{DINO}$ & $+0.596$ & $+0.502$ & $+0.874$ \\
GW$_\textrm{CLIP}$ & $+0.500$ & $+0.896$ & $+0.881$ \\
GW$_\textrm{DINO}$ & $+0.719$ & $+0.760$ & $+0.765$ \\
\bottomrule
\end{tabular}
\end{table}

\begin{table}[t]
\centering
\caption{Signed Spearman $\rho$ between each training-free metric and RT-DETR-l downstream mAP@$0.5{:}0.95$ on COCO \texttt{val2017} ($18$ synthetic pools plus the real oracle, $19$ points). Best in bold.}
\label{tab:rank-rtdetr}
\setlength{\tabcolsep}{5pt}\small
\begin{tabular}{@{}lccc@{}}
\toprule
Metric & Classical & CCDM-shared & CCDM-classaware \\
\midrule
FID & $+0.925$ & $+0.786$ & $-0.079$ \\
KID & $+0.916$ & $+0.951$ & $+0.947$ \\
FCD$_\textrm{CLIP}$ & $+0.970$ & $+0.968$ & $+0.946$ \\
DINO-FD & $+0.612$ & $+0.567$ & $+0.188$ \\
SWD$_\textrm{CLIP}$ & $+0.951$ & $+0.958$ & $\mathbf{+0.974}$ \\
SWD$_\textrm{DINO}$ & $+0.349$ & $+0.521$ & $+0.565$ \\
CMD$_\textrm{CLIP}$ & $+0.935$ & $+0.960$ & $+0.970$ \\
CMD$_\textrm{DINO}$ & $+0.577$ & $+0.565$ & $+0.651$ \\
Wass$_\textrm{CLIP}$ & $+0.193$ & $+0.496$ & $+0.451$ \\
Wass$_\textrm{DINO}$ & $-0.379$ & $-0.472$ & $-0.263$ \\
MMD-RBF$_\textrm{CLIP}$ & $+0.923$ & $+0.886$ & $+0.923$ \\
MMD-RBF$_\textrm{DINO}$ & $+0.532$ & $+0.474$ & $+0.807$ \\
MMD-ms$_\textrm{CLIP}$ & $+0.919$ & $+0.886$ & $+0.928$ \\
MMD-ms$_\textrm{DINO}$ & $+0.556$ & $+0.472$ & $+0.826$ \\
GW$_\textrm{CLIP}$ & $+0.467$ & $+0.902$ & $+0.865$ \\
GW$_\textrm{DINO}$ & $+0.693$ & $+0.733$ & $+0.826$ \\
\bottomrule
\end{tabular}
\end{table}

\section{Conditioning as a paired gain}
\label{sec:supp:paired}

\Cref{tab:syn-ranking} compares metrics one against another, which leaves open whether the gain comes from conditioning or from the choice of base metric and embedding. This section isolates conditioning by fixing the base metric and toggling the conditioning on and off, so the two numbers in each pair differ in nothing else. \Cref{tab:supp-paired} reports the result on five pool sets, averaged over the base metrics that share a kind of estimator. COCO$_{18}$ ranks the 18 synthetic pools alone and COCO$_{20}$ adds the two real candidates. The last two columns are composition-shifted real pool sets drawn from Open Images~\cite{kuznetsova2020open} and NuImages~\cite{caesar2020nuscenes}, built as described in \Cref{sec:supp:compshift-map}.

Conditioning improves 12 of the 16 base metrics on COCO$_{18}$, 13 on COCO$_{20}$, 12 on VisDrone-DET, and 15 and 11 on the two additional datasets, with a mean gain of $+0.06$ to $+0.21$. The exception is systematic rather than random. The Fr\'echet group (which also contains KID) is the only one that does not benefit consistently (it gains only on Open Images). Its three Fr\'echet members must fit a full feature covariance in each stratum ($2048\times2048$ for FID) from a median of $16$ images, so the per-stratum Gaussian fit is unreliable, whereas KID, a kernel estimator, gains on both main benchmarks (\Cref{tab:syn-ranking}). Every other group gains on every pool set. This is why CCDM uses a kernel base rather than a Fr\'echet one.

\begin{table}[t]
\centering
\caption{\textbf{Conditioning is a paired gain, not an embedding effect.} Signed Spearman $\rho$ against downstream mAP without conditioning (\emph{base}) and with it (\emph{+C}, the better of the two stratification policies), averaged over the base metrics that share a kind of estimator. COCO$_{18}$ ranks the 18 synthetic pools alone, COCO$_{20}$ adds the two real candidates, and VisDrone$_{19}$ is \Cref{tab:syn-ranking} unchanged. OpenImg and NuImg are the composition-shifted real pool sets of \Cref{tab:supp-compshift}, without the three unbiased controls, which carry no bias to resolve. Subscripts are candidate counts, and bold marks a gain from conditioning. $n$ is the number of base metrics averaged. Groups: Fr\'echet $+$ KID = FID, KID, FCD$_\textrm{CLIP}$, DINO-FD; MMD = MMD-RBF and MMD-RBF-multiscale on CLIP and DINOv2; OT = SWD and Wasserstein on CLIP and DINOv2; CMD and GW = each on CLIP and DINOv2.}
\label{tab:supp-paired}
\setlength{\tabcolsep}{4pt}
\resizebox{\columnwidth}{!}{%
\begin{tabular}{@{}lccccccccccc@{}}
\toprule
Estimator & $n$ & \multicolumn{2}{c}{COCO$_{18}$} & \multicolumn{2}{c}{COCO$_{20}$} & \multicolumn{2}{c}{VisDrone$_{19}$} & \multicolumn{2}{c}{OpenImg$_{15}$} & \multicolumn{2}{c}{NuImg$_{14}$} \\
\cmidrule(lr){3-4}\cmidrule(lr){5-6}\cmidrule(lr){7-8}\cmidrule(lr){9-10}\cmidrule(lr){11-12}
 &  & base & +C & base & +C & base & +C & base & +C & base & +C \\
\midrule
Fr\'echet $+$ KID & 4 & $+0.80$ & $+0.77$ & $+0.85$ & $+0.83$ & $+0.85$ & $+0.85$ & $+0.47$ & $\mathbf{+0.58}$ & $+0.76$ & $+0.72$ \\
MMD & 4 & $+0.65$ & $\mathbf{+0.87}$ & $+0.75$ & $\mathbf{+0.91}$ & $+0.90$ & $\mathbf{+0.93}$ & $+0.29$ & $\mathbf{+0.47}$ & $+0.56$ & $\mathbf{+0.63}$ \\
OT & 4 & $+0.15$ & $\mathbf{+0.45}$ & $+0.31$ & $\mathbf{+0.46}$ & $+0.78$ & $\mathbf{+0.79}$ & $+0.32$ & $\mathbf{+0.45}$ & $+0.68$ & $\mathbf{+0.75}$ \\
CMD & 2 & $+0.67$ & $\mathbf{+0.75}$ & $+0.76$ & $\mathbf{+0.82}$ & $+0.84$ & $\mathbf{+0.88}$ & $+0.42$ & $\mathbf{+0.64}$ & $+0.74$ & $\mathbf{+0.81}$ \\
GW & 2 & $+0.75$ & $\mathbf{+0.82}$ & $+0.56$ & $\mathbf{+0.82}$ & $-0.43$ & $\mathbf{-0.05}$ & $-0.21$ & $\mathbf{+0.38}$ & $+0.06$ & $\mathbf{+0.72}$ \\
\midrule
\emph{improved} & 16 & \multicolumn{2}{c}{12/16\,($+0.14$)} & \multicolumn{2}{c}{13/16\,($+0.11$)} & \multicolumn{2}{c}{12/16\,($+0.06$)} & \multicolumn{2}{c}{15/16\,($+0.21$)} & \multicolumn{2}{c}{11/16\,($+0.12$)} \\
\bottomrule
\end{tabular}%
}
\end{table}

\section{Composition-shifted pools and downstream mAP}
\label{sec:supp:compshift-map}

\Cref{tab:comp-shift} shows that CCDM responds to composition while appearance metrics stay constant, but there the images are held identical, so only the metric moves. This section closes the loop to mAP: we build pool sets whose composition is corrupted, train one detector per pool, and ask whether the metric ordering matches the mAP ordering.

From a fixed parent pool we draw $3{,}000$-image subsets, each biased along exactly one composition axis, object count, box scale, or class mix, at several strengths, plus three unbiased control subsets drawn from the same parent with no bias applied. Every subset trains one YOLOv8m from scratch under the locked recipe, so the metric score and the mAP in each row describe the same pool. COCO pool sets are built twice, once from synthetic images and once from the index-matched real images, and Open Images and NuImages are real throughout. \Cref{tab:supp-compshift} reports the outcome, with one fixed CCDM variant, class-aware CMD$_\textrm{CLIP}$, used on every pool set rather than the best variant per set.

The controls behave as they should. They attain the highest mAP and the metrics place them at the top, so the ranking is not an artifact of subset size. Biasing any single axis costs mAP, and CCDM reads that loss: it leads FID, SWD$_\textrm{CLIP}$, and GW$_\textrm{CLIP}$ on three of the four pool sets and stays within $0.03$ of the strongest of them on the fourth. Note that subsetting a pool changes appearance as a by-product, so classical metrics are not blind here. The appearance-only control where they do fail is \Cref{tab:comp-shift}.

\begin{table}[t]
\centering
\caption{\textbf{Composition-shifted pools with their downstream mAP.} Each $3{,}000$-image pool set biases one composition axis of a fixed parent pool, six subsets by object count, six by box scale and three by class mix, plus three unbiased controls, and every pool trains one detector under the locked recipe. These sets are unrelated to the 18 generator pools of \Cref{tab:syn-ranking}, and NuImages has 17 because its \emph{drop-top-5} subset leaves only $605$ images, the five most frequent classes appearing in nearly every driving frame. $\rho$ is signed Spearman against test mAP@$0.5$. CCDM is the class-aware CMD$_\textrm{CLIP}$ variant chosen once rather than per dataset, CMD$_\textrm{CLIP}$ is that base unconditioned, and Meta uses metadata only.}
\label{tab:supp-compshift}
\setlength{\tabcolsep}{4pt}
\resizebox{\columnwidth}{!}{%
\begin{tabular}{@{}lcccccccc@{}}
\toprule
Pool set & $n$ & mAP@$0.5$ range & \multicolumn{6}{c}{$\rho$ vs.\ mAP} \\
\cmidrule(lr){4-9}
 &  &  & FID & SWD & GW & CMD$_\textrm{CLIP}$ & Meta & CCDM \\
\midrule
COCO (real)  & 18 & $0.075$--$0.184$ & $+0.83$ & $\mathbf{+0.84}$ & $-0.05$ & $+0.79$ & $+0.79$ & $+0.81$ \\
COCO (synth) & 18 & $0.051$--$0.147$ & $+0.86$ & $+0.84$ & $+0.49$ & $\mathbf{+0.87}$ & $+0.72$ & $\mathbf{+0.87}$ \\
Open Images  & 18 & $0.064$--$0.150$ & $+0.57$ & $+0.43$ & $-0.16$ & $+0.46$ & $\mathbf{+0.67}$ & $+0.61$ \\
NuImages     & 17 & $0.147$--$0.261$ & $+0.77$ & $+0.70$ & $+0.28$ & $+0.75$ & $+0.59$ & $\mathbf{+0.78}$ \\
\bottomrule
\end{tabular}%
}
\end{table}

\section{Sensitivity to the design choices}
\label{sec:supp:sensitivity}

CCDM fixes two things by hand: which metadata axes form the stratum key, and how finely each axis is binned. This section reports what happens when each is varied.

\noindent\textbf{Metadata axes.} We recompute CCDM with subsets of the three axes, averaging over the 16 base metrics (\Cref{tab:supp-loo}). No single axis wins everywhere. On the appearance-varying COCO benchmark the object-count axis alone ranks highest, on the composition-shifted COCO pools the class axis alone does, and on Open Images and NuImages the full three-axis grid does. Since the winning axis differs by dataset and cannot be known before the pools are trained, using all three is the choice that never falls far behind, and it beats the unconditioned base on every pool set, which single-axis subsets do not.

\noindent\textbf{Bin granularity.} We sweep the density$\times$scale grid from $2\times2$ to $6\times7$ (\Cref{tab:supp-bins}). The coarse $2\times2$ grid is best on three of the four $3{,}000$-image pool sets, and NuImages prefers $4\times5$. On the $10{,}000$-image main benchmark $2\times2$ and $3\times4$ are effectively tied ($+0.654$ each). The pattern is a sample-budget effect rather than a universal optimum: finer strata resolve composition better but leave fewer images per stratum, so the useful granularity grows with the number of images available and is not fixed across datasets.

\begin{table}[t]
\centering
\caption{Stratum-key ablation. Signed Spearman $\rho$ against mAP, averaged over the 16 base metrics, for subsets of the three metadata axes. ``base'' is the unconditioned metric. The best axis subset differs by pool set, so the full three-axis key is the default that is never far from the best.}
\label{tab:supp-loo}
\setlength{\tabcolsep}{6pt}
\resizebox{\columnwidth}{!}{%
\begin{tabular}{@{}lcccc@{}}
\toprule
Axis subset & COCO$_{20}$ & COCO$_{\text{shift}}$ & OpenImg & NuImg \\
\midrule
class $+$ count $+$ scale & $+0.673$ & $+0.701$ & $\mathbf{+0.456}$ & $\mathbf{+0.712}$ \\
count $+$ scale           & $+0.671$ & $+0.596$ & $+0.373$ & $+0.620$ \\
class                     & $+0.668$ & $\mathbf{+0.784}$ & $+0.378$ & $+0.574$ \\
count                     & $\mathbf{+0.699}$ & $+0.560$ & $+0.353$ & $+0.510$ \\
scale                     & $+0.575$ & $+0.523$ & $+0.294$ & $+0.524$ \\
\midrule
base (unconditioned)      & $+0.641$ & $+0.578$ & $+0.295$ & $+0.601$ \\
\bottomrule
\end{tabular}%
}
\end{table}

\begin{table}[t]
\centering
\caption{Bin-granularity sweep of the density$\times$scale grid. Signed Spearman $\rho$ against mAP, averaged over the 16 base metrics with shared strata. The first four rows are the $3{,}000$-image composition-shifted pool sets, the last is the $10{,}000$-image main COCO benchmark, reported to three decimals because the two coarsest grids separate only there. Coarse grids win where images per stratum are scarce, and the advantage disappears as the pools grow. The composition-shifted rows exclude the three unbiased controls, so their counts are three lower than in \Cref{tab:supp-compshift}.}
\label{tab:supp-bins}
\setlength{\tabcolsep}{4pt}
\resizebox{\ifdim\width>\columnwidth \columnwidth\else\width\fi}{!}{%
\begin{tabular}{@{}lccccc@{}}
\toprule
Pool set & $n$ & $2\times2$ & $3\times4$ & $4\times5$ & $6\times7$ \\
\midrule
COCO (real)  & 15 & $\mathbf{+0.72}$ & $+0.45$ & $+0.43$ & $+0.45$ \\
COCO (synth) & 15 & $\mathbf{+0.69}$ & $+0.57$ & $+0.55$ & $+0.45$ \\
Open Images  & 15 & $\mathbf{+0.38}$ & $+0.37$ & $+0.30$ & $+0.12$ \\
NuImages     & 14 & $+0.57$ & $+0.62$ & $\mathbf{+0.63}$ & $+0.61$ \\
\midrule
COCO main    & 19 & $\mathbf{+0.654}$ & $\mathbf{+0.654}$ & $+0.646$ & $+0.609$ \\
\bottomrule
\end{tabular}%
}
\end{table}

\section{Pool size and seed variance}
\label{sec:supp:stability}

Two questions bear on whether CCDM is usable in practice: how many images a candidate pool needs before its ranking is trustworthy, and how much the ranking moves when the benchmark itself is rebuilt.

\noindent\textbf{How small can a candidate pool be?} We score the 18 COCO pools from random subsets of increasing size against the fixed $5{,}000$-image reference, averaging over three draws (\Cref{tab:supp-subsample}). Down to $2{,}500$ images, a quarter of the full pool, every CCDM variant on a sample-based base still ranks the pools at $\rho = +0.80$ to $+0.93$ against mAP. Below $1{,}000$ images the estimates degrade, and FID degrades fastest, reversing sign at $250$. This is what makes a generate-then-screen loop practical: one can render a small trial pool per generator configuration, keep the configuration CCDM ranks best, and generate at scale only for that one.

\begin{figure*}[t]
  \centering
  \includegraphics[width=\textwidth]{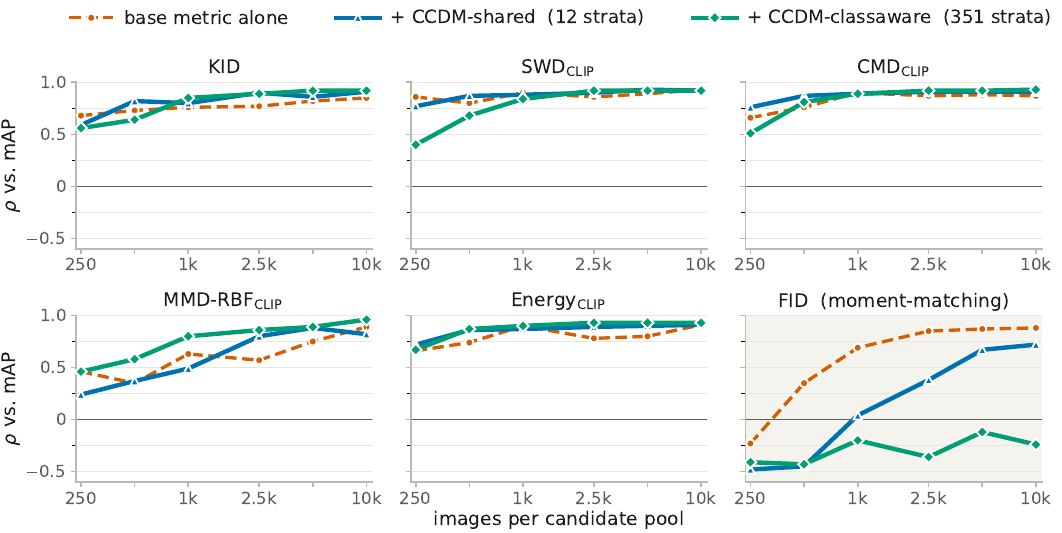}
  \caption{\textbf{How quickly each metric becomes reliable.} Signed Spearman $\rho$ against mAP over the 18 COCO pools as the candidate pool is subsampled, reference fixed at $5{,}000$ images and averaged over three draws. A stratum enters $D_\textrm{cond}$ only when the reference and the candidate each hold at least four images in it, so finer strata empty out first, and at $250$ images the $12$ shared strata still cover $97.5\%$ of the reference mass while the $351$ class-aware strata cover only $40.6\%$. At that size conditioning costs accuracy for KID, SWD$_\textrm{CLIP}$, and FID, and by $2{,}500$ the better conditioning leads the base in all five sample-based panels, by $+0.29$ for MMD-RBF$_\textrm{CLIP}$. FID is the exception at every size, since its per-stratum Gaussian estimate cannot be formed from so few images, the collapse reported in \Cref{tab:syn-ranking}. \Cref{tab:supp-subsample} lists these values together with two further moment-matching bases.}
  \label{fig:subsample-curve}
\end{figure*}

\noindent\textbf{What degrades at small pool sizes.} A stratum contributes to $D_\textrm{cond}$ only when the reference and the candidate each hold at least four images in it, and strata below that floor are skipped, so a pool too small to populate them falls back to the global distance. The shared partition has $12$ strata and still covers $97.5\%$ of the reference mass at $250$ images, while the class-aware partition has $351$ and covers only $40.6\%$, which is the order in which the curves fall away in \Cref{fig:subsample-curve}. Below roughly $500$ images per pool the coarse partition is therefore the one to use.

\noindent\textbf{Does the ranking survive a rebuild?} We rebuilt the composition-shifted COCO synthetic benchmark end to end under three sampling seeds, redrawing every subset and retraining every detector, which gives a distribution over seeds for each metric (\Cref{tab:supp-seed}). The spread is small for the metrics that rank well, CCDM at $+0.847 \pm 0.028$ and FID at $+0.834 \pm 0.026$ (mean and standard deviation over seeds), and CCDM ranks first in all three seeds. GW$_\textrm{CLIP}$ is the least stable ($\pm 0.123$), consistent with it being the weakest metric.

\begin{table}[t]
\centering
\caption{Subsample sweep on the 18 COCO synthetic pools, reference set fixed at $5{,}000$ images and each entry averaged over three random draws. Signed Spearman $\rho$ against test mAP@$0.5$:$0.95$, with the $10$k column the full pool. Each base metric appears unconditioned, conditioned on the $12$ shared strata and conditioned on the $351$ class-aware strata, and the best of the three is in bold. \Cref{fig:subsample-curve} plots these values.}
\label{tab:supp-subsample}
\footnotesize
\setlength{\tabcolsep}{2.5pt}
\begin{tabular*}{\columnwidth}{@{}ll@{\extracolsep{\fill}}cccccc@{}}
\toprule
Base & & $250$ & $500$ & $1$k & $2.5$k & $5$k & $10$k \\
\midrule
\multicolumn{8}{@{}l}{\emph{sample-based base metrics}} \\
KID & none & $\mathbf{+.68}$ & $+.73$ & $+.76$ & $+.77$ & $+.82$ & $+.85$ \\
 & $+$sh. & $+.59$ & $\mathbf{+.82}$ & $+.80$ & $\mathbf{+.90}$ & $+.86$ & $+.91$ \\
 & $+$cls. & $+.56$ & $+.64$ & $\mathbf{+.85}$ & $+.89$ & $\mathbf{+.92}$ & $\mathbf{+.92}$ \\
SWD$_\textrm{CLIP}$ & none & $\mathbf{+.86}$ & $+.80$ & $\mathbf{+.90}$ & $+.86$ & $+.89$ & $\mathbf{+.93}$ \\
 & $+$sh. & $+.77$ & $\mathbf{+.87}$ & $+.88$ & $+.90$ & $\mathbf{+.93}$ & $+.92$ \\
 & $+$cls. & $+.40$ & $+.68$ & $+.84$ & $\mathbf{+.92}$ & $+.92$ & $+.92$ \\
CMD$_\textrm{CLIP}$ & none & $+.66$ & $+.76$ & $\mathbf{+.90}$ & $+.87$ & $+.88$ & $+.87$ \\
 & $+$sh. & $\mathbf{+.76}$ & $\mathbf{+.87}$ & $+.89$ & $+.91$ & $+.91$ & $+.91$ \\
 & $+$cls. & $+.51$ & $+.81$ & $+.89$ & $\mathbf{+.92}$ & $\mathbf{+.92}$ & $\mathbf{+.93}$ \\
MMD-RBF$_\textrm{CLIP}$ & none & $\mathbf{+.46}$ & $+.35$ & $+.63$ & $+.57$ & $+.75$ & $+.89$ \\
 & $+$sh. & $+.24$ & $+.37$ & $+.49$ & $+.80$ & $+.88$ & $+.82$ \\
 & $+$cls. & $\mathbf{+.46}$ & $\mathbf{+.58}$ & $\mathbf{+.80}$ & $\mathbf{+.86}$ & $\mathbf{+.89}$ & $\mathbf{+.96}$ \\
Energy$_\textrm{CLIP}$ & none & $+.66$ & $+.74$ & $\mathbf{+.90}$ & $+.78$ & $+.80$ & $+.91$ \\
 & $+$sh. & $\mathbf{+.72}$ & $+.86$ & $+.87$ & $+.89$ & $+.90$ & $+.91$ \\
 & $+$cls. & $+.67$ & $\mathbf{+.87}$ & $\mathbf{+.90}$ & $\mathbf{+.93}$ & $\mathbf{+.93}$ & $\mathbf{+.93}$ \\
\midrule
\multicolumn{8}{@{}l}{\emph{moment-matching base metrics}} \\
FID & none & $\mathbf{-.23}$ & $\mathbf{+.35}$ & $\mathbf{+.69}$ & $\mathbf{+.85}$ & $\mathbf{+.87}$ & $\mathbf{+.88}$ \\
 & $+$sh. & $-.48$ & $-.45$ & $+.04$ & $+.38$ & $+.67$ & $+.72$ \\
 & $+$cls. & $-.41$ & $-.43$ & $-.20$ & $-.36$ & $-.12$ & $-.24$ \\
DINO-FD & none & $\mathbf{+.21}$ & $\mathbf{+.42}$ & $\mathbf{+.53}$ & $\mathbf{+.57}$ & $\mathbf{+.56}$ & $\mathbf{+.54}$ \\
 & $+$sh. & $-.22$ & $+.13$ & $+.31$ & $+.42$ & $+.48$ & $+.50$ \\
 & $+$cls. & $-.17$ & $-.07$ & $+.08$ & $+.07$ & $+.17$ & $+.06$ \\
FCD$_\textrm{CLIP}$ & none & $\mathbf{+.88}$ & $\mathbf{+.94}$ & $\mathbf{+.92}$ & $\mathbf{+.93}$ & $+.93$ & $\mathbf{+.93}$ \\
 & $+$sh. & $+.58$ & $+.87$ & $+.91$ & $\mathbf{+.93}$ & $\mathbf{+.94}$ & $\mathbf{+.93}$ \\
 & $+$cls. & $+.23$ & $+.50$ & $+.69$ & $+.81$ & $+.89$ & $+.88$ \\
\bottomrule
\end{tabular*}
\end{table}

\begin{table}[t]
\centering
\caption{Seed variance of the composition-shifted COCO synthetic benchmark. For each seed the subsets are redrawn and every detector is retrained from scratch, and $\rho$ is signed Spearman against the resulting mAP over 18 pools. CCDM ranks first in every seed, and its spread over seeds is comparable to FID's.}
\label{tab:supp-seed}
\footnotesize
\setlength{\tabcolsep}{5pt}
\begin{tabular*}{\columnwidth}{@{}l@{\extracolsep{\fill}}cccc@{}}
\toprule
Seed & FID & SWD$_\textrm{CLIP}$ & GW$_\textrm{CLIP}$ & CCDM \\
\midrule
$s_1$ & $+0.864$ & $+0.837$ & $+0.488$ & $\mathbf{+0.874}$ \\
$s_2$ & $+0.816$ & $+0.812$ & $+0.653$ & $\mathbf{+0.818}$ \\
$s_3$ & $+0.820$ & $+0.833$ & $+0.730$ & $\mathbf{+0.849}$ \\
\midrule
mean & $+0.834$ & $+0.827$ & $+0.624$ & $\mathbf{+0.847}$ \\
std  & $\pm0.026$ & $\pm0.013$ & $\pm0.123$ & $\pm0.028$ \\
\bottomrule
\end{tabular*}
\end{table}

\section{Scope and limitations}
\label{sec:supp:limitations}
CCDM selects under a declared target, so it is complementary to domain randomization rather than an alternative to it. When the deployment domain is unknown or very broad, maximizing coverage is the right strategy. CCDM answers a narrower question: which candidate training set $\mathcal{C}$ to train on once a reference target $\mathcal{T}$ is given. Two practical considerations follow from this design, and we close by clarifying how CCDM relates to diversity-maximizing practice.

CCDM is a wrapper over a base feature distance rather than a standalone distance. It partitions $\mathcal{C}$ and $\mathcal{T}$ into metadata strata over object-count and object-scale bins (refined by dominant class in the class-aware variant) and aggregates a base feature distance $D_{\text{base}}$ (FID, KID, MMD-RBF, GW, etc.) computed per stratum. Its behavior therefore tracks that of the wrapped distance: a base distance that is uninformative in the deployment regime stays uninformative after stratification, and per-stratum evaluation sees fewer samples than a global one, which matters for sample-hungry base metrics when the reference set is small. In return, the wrapper design is what lets CCDM upgrade whichever base distance best suits a given feature space and sample budget, rather than committing to a single distance.

CCDM also assumes a reference set $\mathcal{T}$ that represents the deployment distribution well. In practice such a set may be unavailable, small, or biased, for example when collected under convenience sampling, and a biased reference redefines what matching means: candidates are then ranked by adherence to the flawed reference. CCDM measures match to the declared target; auditing the target itself is outside its scope and is standard practice in dataset curation regardless of the selection metric used.

Finally, rewarding adherence may appear to contradict the common advice that diversity benefits object-detection training, with domain randomization as the canonical example \cite{mehta2020active}. The contradiction is only apparent. CCDM constrains the stratum-mass distribution $p^{\mathcal{C}}$, the share of images in each metadata stratum, and does not penalize visual diversity within a stratum; a diverse reference stratum is best matched by equally diverse candidate content. Composition-level adherence and within-stratum diversity are therefore orthogonal axes, and the two prescriptions are composable: match the composition, diversify the content.

\end{document}